\documentclass[10pt,twocolumn,letterpaper]{article}

\usepackage{iccv}
\usepackage{times}
\usepackage{epsfig}
\usepackage{graphicx}
\usepackage{amsmath}
\usepackage{amssymb}
\usepackage{multirow}
\usepackage{booktabs}
\usepackage{tikz}
\usepackage{physics}
\usepackage{upgreek}
\usepackage{gensymb}
\usepackage[accsupp]{axessibility}  
\usepackage{appendix}

\definecolor{gold}{RGB}{221, 196, 65}
\definecolor{silver}{RGB}{215, 215, 215}
\definecolor{bronze}{RGB}{126, 66, 5}
\newcommand{\tikzcircle}[2][red,fill=red]{\tikz[baseline=-0.7ex]\draw[#1,radius=#2] (0,0) circle ;}

\usepackage[pagebackref=true,breaklinks=true,letterpaper=true,colorlinks,bookmarks=false]{hyperref}

\iccvfinalcopy 

\def\iccvPaperID{5223} 

\ificcvfinal\fi

\begin{document}

\title{NeuRBF: A Neural Fields Representation with Adaptive Radial Basis Functions}

\author{Zhang Chen$^{1\dagger}$
\qquad
Zhong Li$^{1\dagger}$
\qquad
Liangchen Song$^{2}$
\qquad
Lele Chen$^{1}$\\
Jingyi Yu$^{3}$
\qquad
Junsong Yuan$^{2}$
\qquad
Yi Xu$^{1}$\\
$^{1}$ OPPO US Research Center \qquad $^{2}$ University at Buffalo \qquad $^{3}$ ShanghaiTech University\\
{\tt\small \{zhang.chen,zhong.li,lele.chen,yi.xu\}@oppo.com}\\
{\tt\small \{lsong8,jsyuan\}@buffalo.edu \qquad yujingyi@shanghaitech.edu.cn}\\
{\tt\small \url{https://oppo-us-research.github.io/NeuRBF-website/}}
}

\maketitle

\ificcvfinal\thispagestyle{empty}\fi

\begin{abstract}
  We present a novel type of neural fields that uses general radial bases for signal representation. State-of-the-art neural fields typically rely on grid-based representations for storing local neural features and N-dimensional linear kernels for interpolating features at continuous query points. The spatial positions of their neural features are fixed on grid nodes and cannot well adapt to target signals. Our method instead builds upon general radial bases with flexible kernel position and shape, which have higher spatial adaptivity and can more closely fit target signals. To further improve the channel-wise capacity of radial basis functions, we propose to compose them with multi-frequency sinusoid functions. This technique extends a radial basis to multiple Fourier radial bases of different frequency bands without requiring extra parameters, facilitating the representation of details. Moreover, by marrying adaptive radial bases with grid-based ones, our hybrid combination inherits both adaptivity and interpolation smoothness. We carefully designed weighting schemes to let radial bases adapt to different types of signals effectively. Our experiments on 2D image and 3D signed distance field representation demonstrate the higher accuracy and compactness of our method than prior arts. When applied to neural radiance field reconstruction, our method achieves state-of-the-art rendering quality, with small model size and comparable training speed. 
\end{abstract}

{\let\thefootnote\relax\footnote{{{$^{\dagger}$} Corresponding author.}}}

\begin{figure}[t]
  \centering
  \includegraphics[width=1.0\columnwidth]{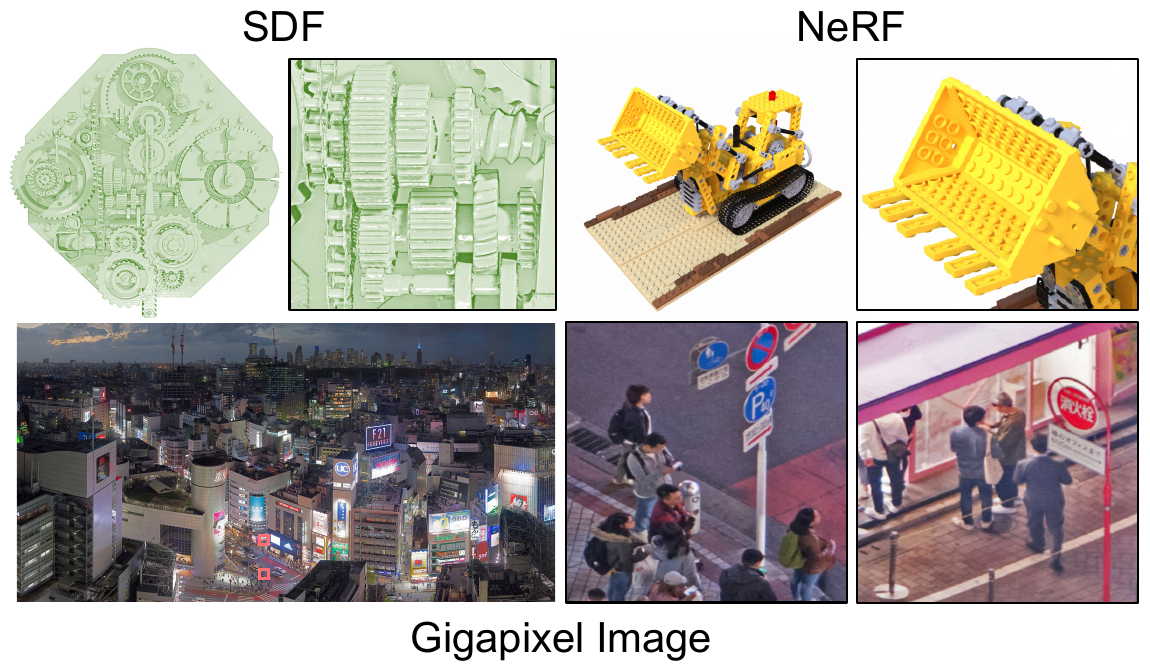}
  \caption{\textbf{NeuRBF} provides an accurate and compact neural fields representation for 2D images, 3D SDF, and neural radiance fields.
  }
  \label{fig:teaser}
\end{figure}

\section{Introduction}
Neural fields (also termed implicit neural representation) have gained much popularity in recent years due to their effectiveness in representing 2D images, 3D shapes, radiance fields, \etc~\cite{park2019deepsdf, mescheder2019occupancy, chen2019learning, tancik2020fourier, sitzmann2020implicit, mildenhall2021nerf, muller2022instant}. Compared to traditional discrete signal representations, neural fields utilize neural networks to establish a mapping from continuous input coordinates to the corresponding output value. Owing to their concise and efficient formulation, neural fields have been applied to various areas ranging from signal compression~\cite{strumpler2022implicit, dupont2022coin++, chen2021nerv, zhang2021implicit}, 3D reconstruction~\cite{oechsle2021unisurf, yariv2021volume, xu2023dynamic}, neural rendering~\cite{mildenhall2021nerf, muller2022instant, chen2022tensorf, li2021neulf, song2022nerfplayer, fridovich2023k, song2023harnessing, chen2020neural}, medical imaging~\cite{feng2022spatiotemporal, yang2022sci, wolterink2022implicit}, acoustic synthesis~\cite{chen2023novel} and climate prediction~\cite{huang2022compressing}.

Early neural fields~\cite{park2019deepsdf, mescheder2019occupancy, chen2019learning, mildenhall2021nerf} use neural features that are globally shared in the input domain. Despite the compactness of the models, they have difficulty in representing high-frequency details due to the inductive bias~\cite{bietti2019inductive, tancik2020fourier} of MLPs. To tackle this problem, local neural fields have been proposed and widely adopted~\cite{chabra2020deep, jiang2020local, peng2020convolutional, liu2020neural, fridovich2022plenoxels, sun2022direct, muller2022instant, chen2022tensorf}, where each local region in the input domain is assigned with different neural features.
A common characteristic in this line of work is to use explicit grid-like structures to spatially organize neural features and apply N-dimensional linear interpolation to aggregate local neural features.
However, grid-like structures are not adaptive to the target signals and cannot fully exploit the non-uniformity and sparsity in various tasks, leading to potentially sub-optimal accuracy and compactness.
While multi-resolution techniques~\cite{takikawa2021neural, chen2021multiresolution, saragadam2022miner, yu2021plenoctrees, han2023multiscale} have been explored, it can still be expensive to achieve fine granularity with excessive resolution levels.
Some works~\cite{mildenhall2021nerf, tancik2020fourier, sitzmann2020implicit} use frequency encoding to address the low-frequency inductive bias. However, this technique is only applied on either input coordinates or deep features.

In this work, we aim to increase the representation accuracy and compactness of neural fields by equipping the interpolation of basis functions with both spatial adaptivity and frequency extension.
We observe that the grid-based linear interpolation, which is the fundamental building block in state-of-the-art local neural fields, is a special case of radial basis function (RBF). 
While grid-based structures typically grow quadratically or cubically, general RBFs can require fewer parameters (sometimes even constant number) to represent patterns such as lines and ellipsoids.
Based upon this observation, we propose NeuRBF, which comprises of a combination of adaptive RBFs and grid-based RBFs. The former uses general anisotropic kernel function with high adaptivity while the latter uses N-dimensional linear kernel function to provide interpolation smoothness. 

To further enhance the representation capability of RBFs, we propose to extend them channel-wise and compose them with multi-frequency sinusoid function. This allows each RBF to encode a wider range of frequencies without requiring extra parameters. 
This multi-frequency encoding technique is also applicable to the features in the MLP, which further improves accuracy and compactness.

To effectively adapt radial bases to target signals, we adopt the weighted variant of K-Means to initialize their kernel parameters, and design a weighting scheme for each of the three tasks (see Fig.~\ref{fig:teaser}): 2D image fitting, 3D signed distance field (SDF) reconstruction, and neural radiance field (NeRF) reconstruction. For NeRF, since it involves indirect supervision, traditional K-Means cannot be directly applied. To address this, we further propose a distillation-based approach.

In summary, our work has the following contributions:
\begin{itemize}
\item We present a general framework for neural fields based on radial basis functions and propose a hybrid combination of adaptive RBFs and grid-based RBFs. 
\item We extend radial bases with multi-frequency sinusoidal composition, which substantially enhances their representation ability.
\item To effectively adapt RBFs to different target signals, we devise tailored weighting schemes for K-Means and a distillation-based approach.
\item Extensive experiments demonstrate that our method achieves state-of-the-art accuracy and compactness on 2D image fitting, 3D signed distance field reconstruction, and neural radiance field reconstruction.
\end{itemize}

\section{Related Work}

\paragraph{Global Neural Fields.}
Early neural fields~\cite{park2019deepsdf, mescheder2019occupancy, chen2019learning, xu2019disn, michalkiewicz2019deep, duan2020curriculum} are global ones and primarily focus on representing the signed distance field (SDF) of 3D shapes. They directly use spatial coordinates as input to multi-layer perceptrons (MLPs) and optionally concatenate a global latent vector for generalized or generative learning. These methods have concise formulation and demonstrate superior flexibility over convolutional neural networks (CNN) and traditional discrete representations in modeling signals in the continuous domain. However, these methods are unable to preserve the high-frequency details in target signals. 

Mildenhall \etal~\cite{mildenhall2021nerf} pioneeringly proposed NeRF, which incorporates neural fields with volumetric rendering for novel view synthesis. They further apply sine transform to the input coordinates (\ie, positional encoding), enabling neural fields to better represent high-frequency details. Similar ideas are also adopted in RFF~\cite{tancik2020fourier} and SIREN~\cite{sitzmann2020implicit}, which use random Fourier features or periodic activation as frequency encoding. These works also promote neural fields to be a general neural representation applicable to different types of signals and tasks. More recently, other encoding functions or architectures have been explored~\cite{fathony2021multiplicative, wang2021spline, lindell2022bacon, shekarforoush2022residual, williams2021neural, williams2022neural, cho2022streamable, landgraf2022pins, zheng2021rethinking, ramasinghe2021learning, ramasinghe2022beyond, chng2022gaussian, yuce2022structured, saragadam2023wire, yang2023polynomial}. For example, MFN~\cite{fathony2021multiplicative} replaces MLPs with the multiplication of multiple linear functions of Fourier or Gabor basis functions, and WIRE~\cite{saragadam2023wire} uses Gabor wavelet as activation function in MLPs.
Radial basis functions (RBF) have also been discussed in~\cite{ramasinghe2021learning, ramasinghe2022beyond}. However, unlike our work, they only consider simplified forms of RBFs and do not explore spatial adaptivity, leading to nonideal performance.

\begin{figure*}
  \centering
  \includegraphics[width=1.0\textwidth]{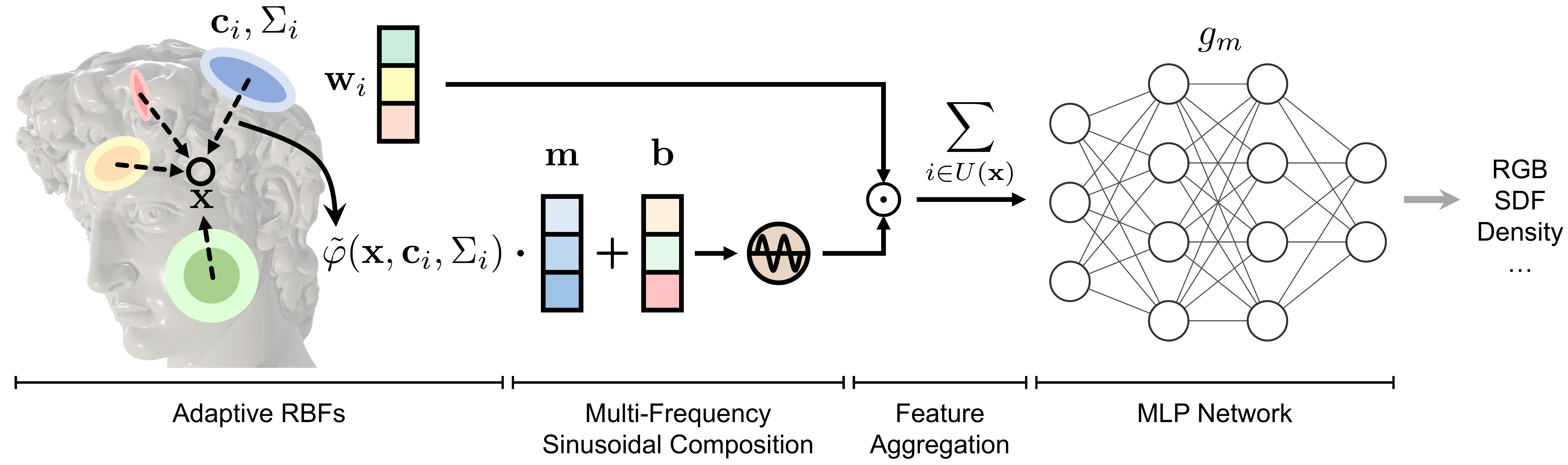}
  \caption{\textbf{Illustration of NeuRBF.} Each adaptive RBFs can have different position and shape parameters $\vb{c}_i, \Sigma_i$, leading to high spatial adaptivity. With multi-frequency sinusoidal composition, each adaptive RBF is further extended to multiple radial bases with different frequencies, which are then combined with neural features $\vb{w}_i$ through Hadamard product. The resulting features are aggregated within the neighborhood $U(\vb{x})$ of query point $\vb{x}$, and then be mapped to the output domain by the MLP network $g_m$.}
  \label{fig:pipeline}
\end{figure*}

\paragraph{Local Neural Fields.}
In parallel to frequency encoding, local neural fields improve representation accuracy by locality. Early attempts~\cite{chabra2020deep, jiang2020local, peng2020convolutional, chibane2020implicit, chen2021learning, sun2022direct} uniformly subdivide the input domain into dense grids and assign a neural feature vector to each grid cell. During point querying, these local neural features are aggregated through nearest-neighbor or N-dimensional linear interpolation and then used as input to the following MLPs. Due to feature locality, the depth and width of the MLPs can be largely reduced~\cite{sun2022direct, fridovich2022plenoxels, karnewar2022relu}, leading to higher training and inference speed than global neural fields. Apart from neural features, the locality can also be implemented on the network weights and biases~\cite{reiser2021kilonerf, saragadam2022miner, haoimplicit}, where each grid cell has a different MLP. 
Dense grids can be further combined with RFF~\cite{tancik2020fourier} or SIREN~\cite{sitzmann2020implicit} to improve accuracy on high-frequency details~\cite{hertz2021sape, mehta2021modulated}. However, a significant drawback of dense grids is that they are parameter-intensive. 

To improve model compactness, numerous techniques have been proposed, such as multi-resolution tree (and/or residual) structures~\cite{liu2020neural, yu2021plenoctrees, chen2021multiresolution, martel2021acorn, saragadam2022miner, yang2022tinc, wu2022neural, fridovich2022plenoxels}, hash grids~\cite{muller2022instant}, dictionary optimization~\cite{takikawa2022variable}, permutohedral lattices~\cite{rosu2023permutosdf}, tensor decomposition~\cite{chen2022tensorf}, orthogonal planes~\cite{peng2020convolutional, chan2022efficient, shue20223d, cao2023hexplane, fridovich2023k}, wavelet~\cite{rho2022masked}, and multiplicative fields composition~\cite{chen2023factor}. Among them, Instant NGP~\cite{muller2022instant} achieves high accuracy, compactness, and efficiency across different signal types. 
Despite the additional data structures or operations, these methods still rely on basic grid-based linear interpolation as the building block for neural feature aggregation. 
Another line of work~\cite{genova2020local, li2022learning, xu2022point} relaxes the grid structures and allows neural features to be freely positioned in the input domain. However, they use simple interpolation kernel functions, which still have limited spatial adaptivity. Their performance is also inferior to state-of-the-art grid-based ones.

Unlike prior local neural fields, we seek a general framework consisting of hybrid radial bases and enhance their representation capability by simultaneously exploiting spatial adaptivity and frequency extension.

\section{Our Method}
\subsection{Local Neural Fields As Radial Basis Functions}
Local neural fields represent a signal $f$ in the form of a function $\hat{f}: \mathbb{R}^D\to\mathbb{R}^O$, which maps a coordinate $\vb{x}$ in the continuous $D$-dimensional space to an output of $O$ dimensions. 
The function $f$ can be considered as a composition of two stages, \ie, $f = g_m \circ g_b$, where $g_b$ extracts the local neural features at input location $\vb{x}$ from a neural representation (\eg, feature grid), and $g_m$ decodes the resulting feature to the final output. 
Now we consider grid-based linear interpolation for $g_b$, which is a common building block in state-of-the-art neural fields. It has the following form: $g_b(\vb{x}) = \sum_{i \in U(\vb{x})} \varphi(\vb{x}, \vb{c}_i) \vb{w}_i$.
$U(\vb{x})$ is the set of grid corner nodes that enclose $\vb{x}$, $\vb{c}_{i} \in \mathbb{R}^D$ and $\vb{w}_i \in \mathbb{R}^F$ are the position and neural feature of node $i$. $\varphi(\vb{x}, \vb{c}_i) \in \mathbb{R}$ is the interpolation weight of node $i$, and is computed as:
\begin{equation}
\varphi(\vb{x}, \vb{c}_i) = \prod_{j=1}^{D} \max(0, 1 - \frac{|\vb{x}_j - \vb{c}_{i,j}|}{\sigma}),
\label{eq:linear_interp}
\end{equation}
where $\sigma$ is the sidelength of each grid cell, and $\vb{x}_j, \vb{c}_{i,j}$ are the $j$th element of $\vb{x}, \vb{c}_{i}$. Note that Eq.~\eqref{eq:linear_interp} is a special case of radial basis function (RBF) with the form of $\varphi(\vb{x}, \vb{c}_i, \sigma_i)$, where each RBF has its own position parameter $\vb{c_i}$ and shape parameter $\sigma_i$.
From the perspective of RBF, we use the following formulation for $g_b(\vb{x})$:
\begin{equation}
g_b(\vb{x}) = \sum_{i \in U(\vb{x})} \varphi(\vb{x}, \vb{c}_i, \sigma_i) \vb{w}_i.
\label{eq:rbf_fields}
\end{equation}

\subsection{Neural Radial Basis Fields}
\begin{figure}
  \centering
  \includegraphics[width=1.0\columnwidth]{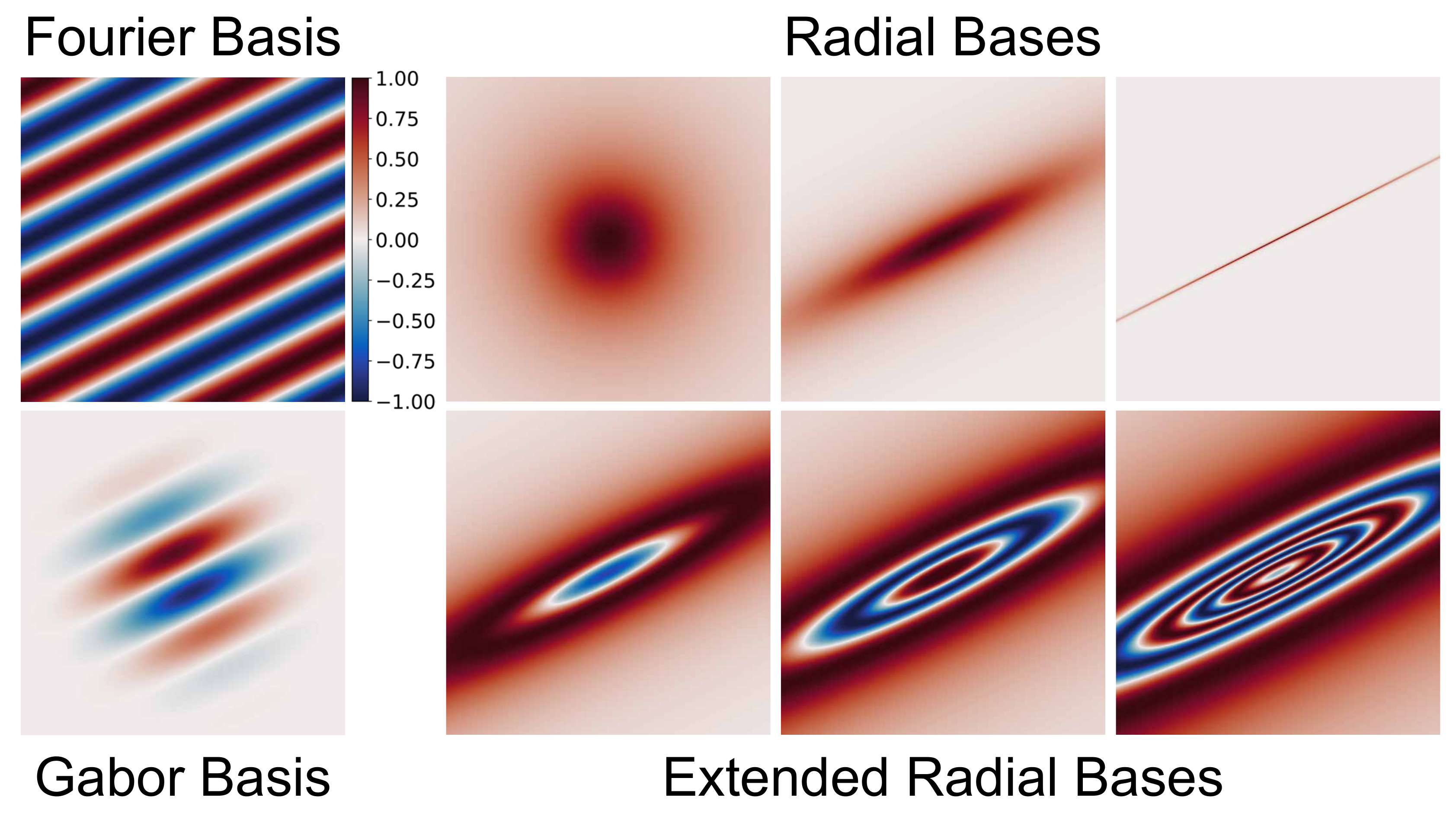}
  \caption{\textbf{Comparison of Bases.} 
  For the right 3 columns: the first row shows radial bases with different shape parameters; the bottom row shows extended radial bases with different frequencies.}
  \label{fig:rbf_illustration}
\end{figure}

Compared to grid-based linear interpolation, the advantages of RBFs originate from the additional position and shape parameters $\vb{c}_i, \sigma_i$. 
As illustrated in Fig.~\ref{fig:pipeline}, our framework makes extensive use of adaptive RBFs.
To fully exploit their adaptivity, we propose to use anisotropic shape parameters $\Sigma_i \in \mathbb{R}^{D \times D}$.
The first row of Fig.~\ref{fig:rbf_illustration} shows that with anisotropic shape parameters, the shape of an RBF's level set can be either circular, elliptical, or even close to a line. This allows an RBF to be more adaptive to target signals. 
For the kernel function $\varphi$, we use the inverse quadratic function as an example, which is computed as:
\begin{equation}
\varphi(\vb{x}, \vb{c}_i, \Sigma_i) = \frac{1}{1 + (\vb{x} - \vb{c}_i)^T\Sigma_i^{-1}(\vb{x} - \vb{c}_i)}.
\label{eq:ivq_a}
\end{equation}
Note that $\Sigma_i$ is a covariance matrix, which is symmetric. Hence, each $\Sigma_i$ only has $\frac{D\cdot(D-1)}{2}$ parameters.
We can optionally normalize the radial basis value at each point:
\begin{equation}
\tilde{\varphi}(\vb{x}, \vb{c}_i, \Sigma_i) = \frac{\varphi(\vb{x}, \vb{c}_i, \Sigma_i)}{\sum_{k \in U(\vb{x})}\varphi(\vb{x}, \vb{c}_k, \Sigma_k)}.
\label{eq:ivq_aa}
\end{equation}

Note that our framework is not limited to a specific function type but supports any others that have the radial basis form. 
The choice of the function type can thus be finetuned per task.

\paragraph{Sinusoidal Composition on Radial Basis.}
We notice that while traditional RBF is a scalar function, $\vb{w}_i\in\mathbb{R}^F$ is a vector with multiple channels (recall Eq.~\eqref{eq:rbf_fields}).
Our motivation is to let each channel of $\vb{w}_i$ linearly combine with a different variant of the RBF so that the channel-wise capacity of RBF can be exploited.
To achieve this, we propose to compose RBF with a multi-frequency sinusoid function, where a radial basis is extended into multiple channels with different frequencies:
\begin{equation}
\boldsymbol{\upvarphi}(\vb{x}, \vb{c}_i, \Sigma_i) = \sin(\tilde{\varphi}(\vb{x}, \vb{c}_i, \Sigma_i) \cdot \vb{m} + \vb{b}),
\label{eq:sin_composition_rbf}
\end{equation}
where $\vb{m}, \vb{b}\in\mathbb{R}^F$ are the global multiplier and bias applied to $\tilde{\varphi}(\vb{x}, \vb{c}_i, \Sigma_i)$ before sine transform. The resulting $\boldsymbol{\upvarphi}(\vb{x}, \vb{c}_i, \Sigma_i)$ has $F$ channels and is then multiplied with $\vb{w}_i$ through Hadamard product. Fig.~\ref{fig:pipeline} illustrates this computation process. $g_b(\vb{x})$ is thus computed as:
\begin{equation}
g_b(\vb{x}) = \sum_{i \in U(\vb{x})} \boldsymbol{\upvarphi}(\vb{x}, \vb{c}_i, \Sigma_i) \odot \vb{w}_i.
\label{eq:rbf_fields_sin}
\end{equation}

With Eq.~\eqref{eq:sin_composition_rbf}, the number of bases encoded by a single pair of $\vb{c}_i, \Sigma_i$ is increased from $1$ to $F$, leading to higher representation ability. 
Note that $\vb{m}, \vb{b}$ are globally shared across RBFs. We set $\vb{b}$ as a learnable parameter and $\vb{m}$ as a fixed parameter. 
We determine the value of $\vb{m}$ by specifying the lowest and highest frequencies $m_l, m_h$. The rest of the elements are obtained by log-linearly dividing the range between $m_l$ and $m_h$.

Our sinusoidal composition technique differs from positional encoding~\cite{mildenhall2021nerf} and random Fourier features~\cite{tancik2020fourier} in that we apply sine transform to radial bases instead of input coordinates. This allows the composited bases to have elliptical periodic patterns as shown in Fig.~\ref{fig:rbf_illustration} second row, while the bases created by ~\cite{mildenhall2021nerf, tancik2020fourier} are limited to linear periodic patterns. 
Our technique is also related to the Gabor filter, which combines a Gaussian function and a sinusoidal function using multiplication. Still, the Gabor filter can only produce bases with linear patterns.

\paragraph{Sinusoidal Composition on Feature Vector.}
We also apply our sinusoidal composition technique to the output features $\vb{h}_0$ of the first fully-connected (FC) layer in $g_m$:
\begin{equation}
\vb{f}_0 = \sin(\vb{h}_0 \odot \vb{m}_0) + \vb{h}_0,
\label{eq:sin_composition_feature}
\end{equation}
where $\vb{h}_0, \vb{m}_0, \vb{f}_0 \in \mathbb{R}^{F_0}$ and $\odot$ is Hadamard product. 
The bias term is omitted since it is already contained in FC layer.
The reason to apply this sinusoidal composition to $\vb{h}_0$ instead of $g_b(\vb{x})$ is to let the network first combines the multi-frequency bases in $g_b(\vb{x})$ via an FC layer. 
Here, we also include a residual connection, which slightly benefits performance. 
The resulting feature vector $\vb{f}_0$ is input to the next layer in $g_m$. 
$\vb{m}_0$ is set in a similar manner as $\vb{m}$ by specifying its lowest and highest frequency $m_{l0}$ and $m_{h0}$.

Compared to sinusoid activation~\cite{sitzmann2020implicit}, our multi-frequency approach can produce features of wide frequency range with one sine transform. In addition, it does not require specialized initialization for the FC layers. We experimentally observe that our technique achieves higher performance under radial basis framework. Table~\ref{tab:ablation_arch} shows a quantitative comparison with positional encoding~\cite{mildenhall2021nerf} and sinusoid activation~\cite{sitzmann2020implicit}.

\paragraph{Hybrid Radial Bases.}
To balance between fitting accuracy and interpolation smoothness, we propose to use a combination of adaptive RBFs and grid-based RBFs. 
The position and shape parameters of adaptive RBFs can be freely assigned while those of grid-based RBFs are fixed to a grid structure. 
Adaptive RBFs tend to produce sharp discontinuities when $U(\vb{x})$ (the set of neighboring RBFs of the point $\vb{x}$) changes. 
On the other hand, grid-based RBFs do not exhibit such discontinuity and can better preserve function smoothness.
Please refer to our supplementary for an illustration.
We combine adaptive and grid-based RBFs through feature concatenation, which allows the network to select features accordingly.

\subsection{Initialization of Position and Shape Parameters}

Motivated by~\cite{schwenker2001three}, we adapt RBFs to target signals by initializing their position and shape parameters with weighted K-Means clustering. Intuitively, this biases RBF distribution towards data points with higher weights. This technique is simple and effective, and can be applied to different tasks by changing the weighting scheme.

\paragraph{Position Initialization.}
Let ${\vb{x}_1, ... , \vb{x}_m}$ be the coordinates of input points and ${w_1, ... , w_m}$ be the weight of each point (weight calculation will be described later). Given initial cluster centers ${\vb{c}_1, ... , \vb{c}_n}$, weighted K-Means optimizes these cluster centers with:
\begin{equation}
\min_{\vb{c}_1, ... , \vb{c}_n} \sum_{i=1}^{n}\sum_{j=1}^{m} a_{ij} w_j \lVert \vb{x}_j - \vb{c}_i \rVert^2,
\label{eq:kmeans}
\end{equation}
where $a_{ij}$ is an indicator variable.
Following common practice, we solve Eq.~\eqref{eq:kmeans} with an expectation–maximization (EM)-style algorithm.

\paragraph{Shape Initialization.}
Inspired by Gaussian mixture model, we initialize the shape parameters $\Sigma_i$ as the following:
\begin{equation}
\Sigma_i = \frac{\sum_j a_{ij} w_j (\vb{x}_j - \vb{c}_i) (\vb{x}_j - \vb{c}_i)^T}{\sum_j a_{ij} w_j}.
\label{eq:init_sigma}
\end{equation}

\paragraph{Weighting Schemes.}
The weights ${w_1, ... , w_m}$ control how RBFs will be distributed after initialization. 
Data points with higher importance should be assigned with higher weights. 

For 2D images, we use the spatial gradient norm of pixel value as the weight for each point: $w_j = \lVert\nabla I(\vb{x}_j)\rVert$. 

For 3D signed distance field, we use the inverse of absolute SDF value as point weight: $w_j = 1\text{ }/\text{ }(|SDF(\vb{x}_j)| + 1e-9)$. The inclusion of $1e-9$ is to avoid division by zero.

For neural radiance field, it is a task with indirect supervision. The input signal is a set of multi-view 2D images while the signal to be reconstructed lies in 3D space. Therefore, we cannot directly obtain the weights.
To tackle this problem, we propose a distillation method. 
We first use grid-based neural fields to train a model for $1000\sim2000$ training steps. 
Then we uniformly sample 3D points and use the trained model to predict the density $\sigma(\vb{x})$ and color feature vector $\vb{f}_c(\vb{x})$ 
at these points. 
Finally, we convert density to alpha and multiply with the spatial gradient norm of the color feature vector as point weight: $w_j = (1 - \text{exp}(-\sigma(\vb{x}_j)\delta))\lVert\nabla \vb{f}_c(\vb{x}_j)\rVert$. 
This weighting scheme takes both density and appearance complexity into account. 
Compared to 3D Gaussian Splatting~\cite{kerbl20233d} and Point-NeRF~\cite{xu2022point}, our approach does not require external structure-from-motion or multi-view stereo methods to reconstruct the point cloud, but distills information from a volumetric model. Hence, our initialization can handle both surface-based objects and volumetric objects.

\section{Implementation}
In this section, we describe the keypoints of our implementation. More details can be found in our supplementary.

We implement our adaptive RBFs using vanilla PyTorch without custom CUDA kernels. For the grid-based part in our framework, we adopt Instant NGP~\cite{muller2022instant} for 2D image fitting and 3D signed distance field (SDF). We use a PyTorch implementation of Instant NGP from~\cite{torchngp}. For neural radiance field (NeRF) reconstruction, we explored TensoRF~\cite{chen2022tensorf} and K-Planes~\cite{fridovich2023k} as the grid-based part.
We reduce the spatial resolution and feature channel of the grid-based part, and allocate parameters to the adaptive RBFs accordingly. 

For sinusoidal composition, we use $m_l=2^{-3}, m_h=2^{12}, m_{l0}=1, m_{h0}=1000$ in the image experiments on DIV2K dataset~\cite{Agustsson_2017_CVPR_Workshops, Timofte_2017_CVPR_Workshops}, and $m_l=2^{0}, m_h=2^{3}, m_{l0}=30, m_{h0}=300$ in SDF experiments. In NeRF task, we do not apply sinusoidal composition since the improvement is small.

Training is conducted on a single NVIDIA RTX A6000 GPU. We use Adam optimizer~\cite{kingma2014adam} where $\beta_1=0.9, \beta_2=0.99, \epsilon=10^{-15}$. 
The learning rates for neural features are $5\times10^{-3}, 1\times10^{-4}, 2\times10^{-2}$ for image, SDF and NeRF task respectively.
In addition, we apply learning rate schedulers that gradually decrease learning rates during training. 
The position and shape parameters of RBFs can be optionally finetuned via gradient backpropagation. However, we do not observe significant performance gain and therefore fix these parameters during training.

We use L2 loss when fitting 2D images and reconstructing neural radiance field, and use MAPE loss~\cite{muller2022instant} when reconstructing 3D SDF. For SDF task, we use the same point sampling approach as Instant NGP~\cite{muller2022instant}. For NeRF task, we follow the training approaches in TensoRF~\cite{chen2022tensorf} and K-Planes~\cite{fridovich2023k} respectively.
In all experiments, both competing methods and our method are trained per scene.

\section{Experiment}\label{sec:exp}

\subsection{2D Image Fitting}
We first evaluate the effectiveness of fitting 2D images. We use the validation split of DIV2K dataset~\cite{Agustsson_2017_CVPR_Workshops, Timofte_2017_CVPR_Workshops} and $6$ additional images of ultra-high resolution as evaluation benchmark. 
DIV2K validation set contains $100$ natural images with resolution around $2040\times1356$. The resolution of the $6$ additional images ranges from $6114\times3734$ to $56718\times21450$.

We first compare with MINER~\cite{saragadam2022miner} and Instant NGP (``I-NGP")~\cite{muller2022instant}, which exhibit state-of-the-art performance for high-resolution image fitting. 
We let our method use fewer parameters than the other two. During timing, the time for initializing RBFs is also taken into account.

\begin{table}[t]
\begin{center}
\resizebox*{\columnwidth}{!}{
\begin{tabular}{lccc|c}
\toprule
 & Steps & Time$\downarrow$ & \# Tr. Params$\downarrow$ & PSNR$\uparrow$ \\
\midrule
\multicolumn{5}{c}{DIV2K} \\
\midrule
MINER~\cite{saragadam2022miner} & 35k & 16.7m & 5.49M & 46.92 \\
I-NGP~\cite{muller2022instant} & 35k & 1.3m & 4.91M & 47.56 \\
Ours & 35k & 7.9m & 4.31M & \textbf{58.56} \\
$\text{Ours}_{3.5k-steps}$ & 3.5k & \textbf{48s} & 4.31M & 51.53 \\
$\text{Ours}_{2.2M}$ & 35k & 7.7m & \textbf{2.20M} & 49.26 \\
\midrule
\multicolumn{5}{c}{DIV2K 256$\times$256$\times$3} \\
\midrule
BACON~\cite{lindell2022bacon} & 5k & 78.2s & 268K & 38.51 \\
PNF~\cite{yang2023polynomial} & 5k & 483.9s & 287K & 38.99 \\
Ours & 5k & \textbf{28.5s} & \textbf{128K} & \textbf{54.84} \\
\bottomrule
\end{tabular}
}
\end{center}
\caption{\textbf{2D Image Fitting.} We quantitatively compare our method with MINER~\cite{saragadam2022miner}, Instant NGP (``I-NGP")~\cite{muller2022instant}, BACON~\cite{lindell2022bacon} and PNF~\cite{yang2023polynomial} on the validation set of DIV2K dataset~\cite{Agustsson_2017_CVPR_Workshops, Timofte_2017_CVPR_Workshops}. ``DIV2K": original image resolution; ``DIV2K 256$\times$256$\times$3": center cropped and downsampled to 256$\times$256$\times$3.
}
\label{tab:image_comp}
\end{table}

\begin{figure}[t]
  \centering
  \includegraphics[width=1.0\columnwidth]{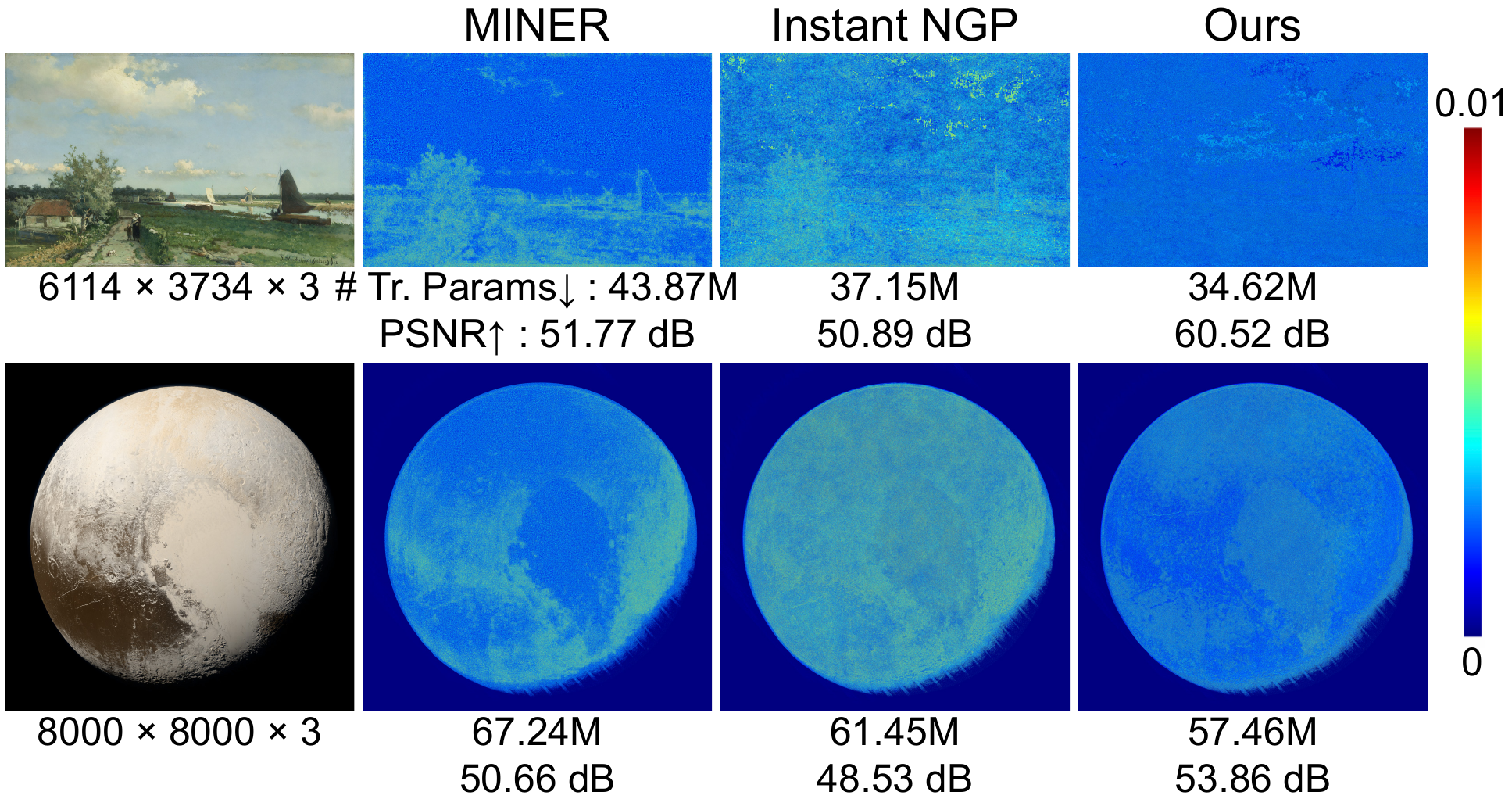}
  \caption{\textbf{2D Image Fitting.} Leftmost column shows the fitted images of our method and the resolution of the images. The other columns show the error maps of each method, along with the number of trainable parameters (``\# Tr. Params") and PSNR.}
  \label{fig:img_comp}
\end{figure}

\begin{figure}[t]
  \centering
  \includegraphics[width=0.985\columnwidth]{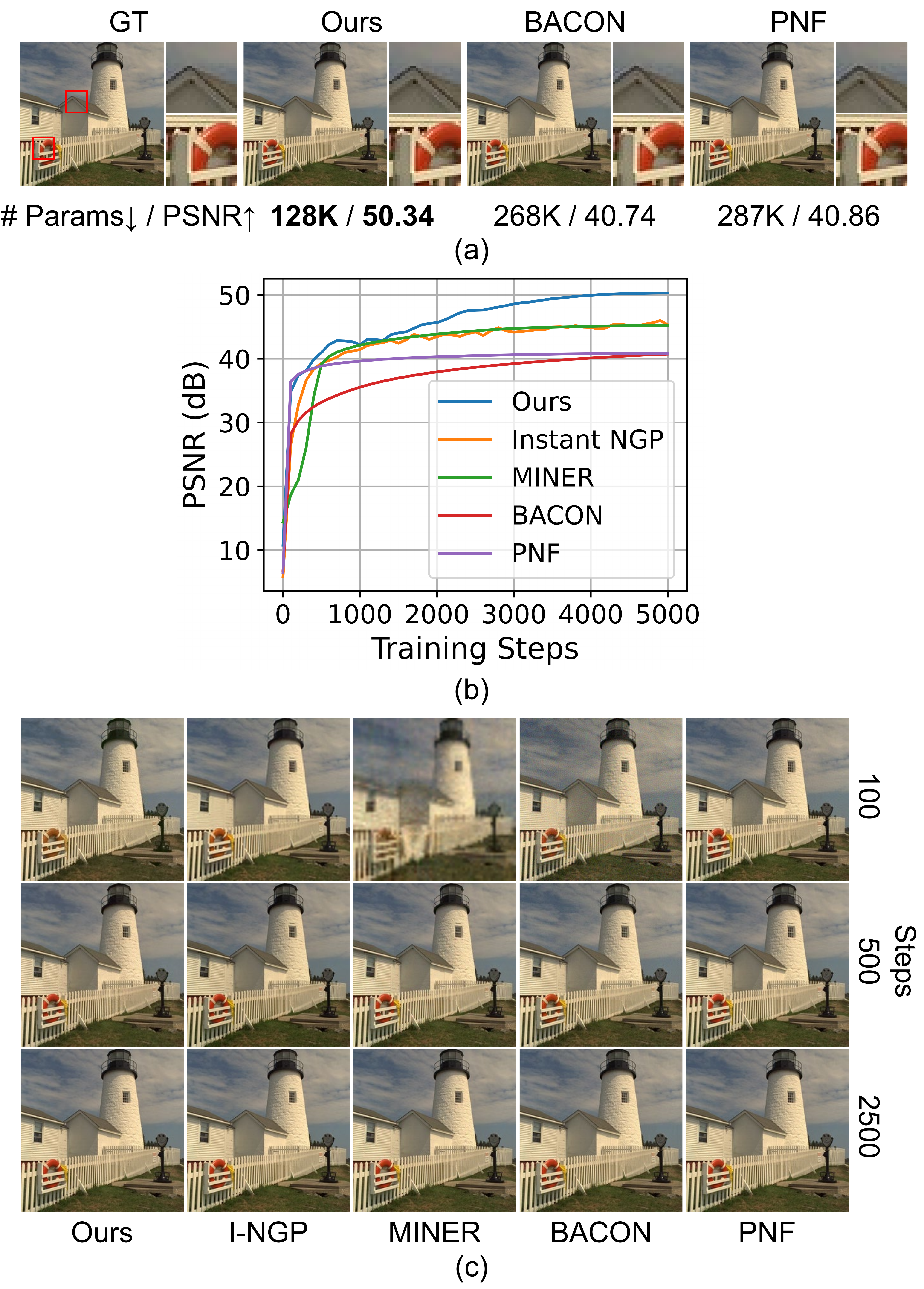}
  \caption{\textbf{2D Image Fitting} on an image from Kodak dataset~\cite{kodak}. (a) Final results after 5k training steps. (b) Training curves. (c) Intermediate results. 
  }
  \label{fig:image_training_curve}
\end{figure}

Table~\ref{tab:image_comp} top half shows the comparison on the DIV2K dataset. For our method, we include two additional setups: one using fewer training steps and one using fewer trainable parameters. 
When using the same number of training steps, our method outperforms the other two by over $10$ dB in Peak Signal-to-Noise Ratio (PSNR) with less trainable parameters. 
Although Instant NGP has faster training speed due to their heavily-optimized CUDA implementation, our method is implemented with vanilla PyTorch and is easily extensible. 
In addition, with only 3.5k training steps ($1/10$ of that of Instant NGP), our method already reaches a PSNR of $51.53$ dB, which is almost $4$ dB higher than Instant NGP. 
Meanwhile, the training time is only 48s and even faster than Instant NGP. The time for RBF initialization is around 2s. ``Ours$_{2.2M}$" additionally demonstrates the high compactness of our method. 
After reducing trainable parameters to be over 50\% fewer than the competing methods, our approach still retains a higher fitting accuracy. 

In Fig.~\ref{fig:img_comp}, we show the fitting results on 2 ultra-high resolution images. Besides achieving higher PSNR than the other two, our method also has a more uniform error distribution. This reflects the adaptivity of RBFs, which allows a more accurate representation of details. Results on other images can be found in our supplementary material. 

We additionally compare with BACON~\cite{lindell2022bacon} and PNF~\cite{yang2023polynomial} on the 100 images in DIV2K validation set. In this experiment, the images are center cropped and downsampled to 256$\times$256$\times$3 following the practice of BACON~\cite{lindell2022bacon}. We use their official codes and settings for BACON and PNF, and let our method use the same batch size (65,536) and training steps (5k) as them. The results are shown in Table~\ref{tab:image_comp} bottom half. 
We further conduct comparisons on a sample image from Kodak dataset~\cite{kodak}, and show the qualitative results and training curves in Fig.~\ref{fig:image_training_curve}. 
The image is similarly center cropped and resized to 256$\times$256$\times$3.
The results show that our method has both fast convergence and high fitting accuracy. 
Higher PSNR demonstrates the ability to more precisely represent target signals, and implies fewer parameters and training steps to reach a specified PSNR. For the image in Fig.~\ref{fig:image_training_curve}, Instant NGP and MINER reach $45.34$ dB and $45.23$ dB PSNR with 140K parameters and 5k steps. Our method instead can reach $45.59$ dB PSNR with only 72K parameters and 3.5k steps.

\begin{figure*}[t]
  \centering
  \includegraphics[width=1.0\textwidth]{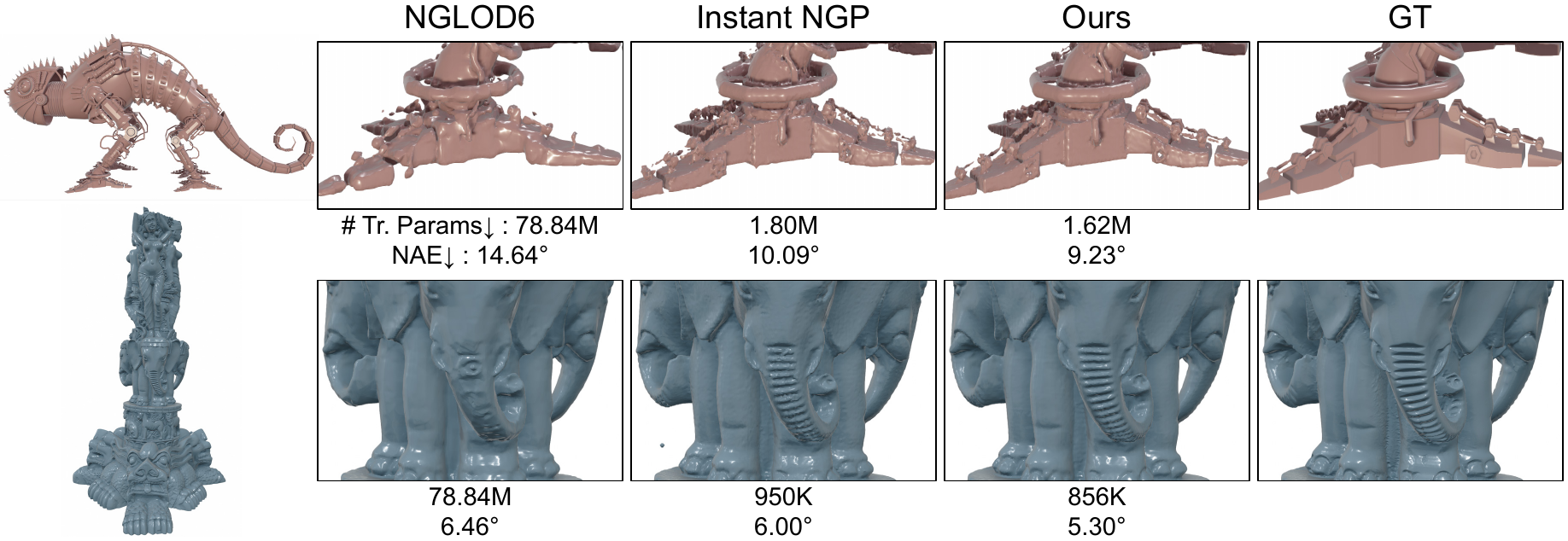}
  \caption{\textbf{3D Signed Distance Field Reconstruction.} Leftmost column shows the reconstructed geometry of our method. The other columns show qualitative and quantitative comparisons of reconstruction results. ``\# Tr. Params" is the number of trainable parameters and ``NAE" is the normal angular error.}
  \label{fig:sdf_comp}
\end{figure*}

\subsection{3D Signed Distance Field Reconstruction}\label{sec:exp_sdf}
We use 10 3D models from the Stanford 3D Scanning Repository~\cite{stanford3d}, the Digital Michelangelo Project~\cite{levoy2000digital}, and TurboSquid~\cite{turbosquid} as benchmark data. These models contain delicate geometric details and challenging topologies. We compare our method with NGLOD~\cite{takikawa2021neural} and Instant NGP~\cite{muller2022instant}. For evaluation metrics, we use Intersection over Union (IoU) and normal angular error (NAE). 
NAE measures the face normal difference of corresponding points and can better reflect the accuracy of reconstructed surface than IoU.

Fig.~\ref{fig:sdf_comp} demonstrates example results on 3 objects. Our method produces more accurate geometry, with sharp edges and smooth surfaces. 
Comparatively, the results of NGLOD are overly smooth while those of Instant NGP contain noises.

\begin{figure}[t]
  \centering
  \includegraphics[width=1\columnwidth]{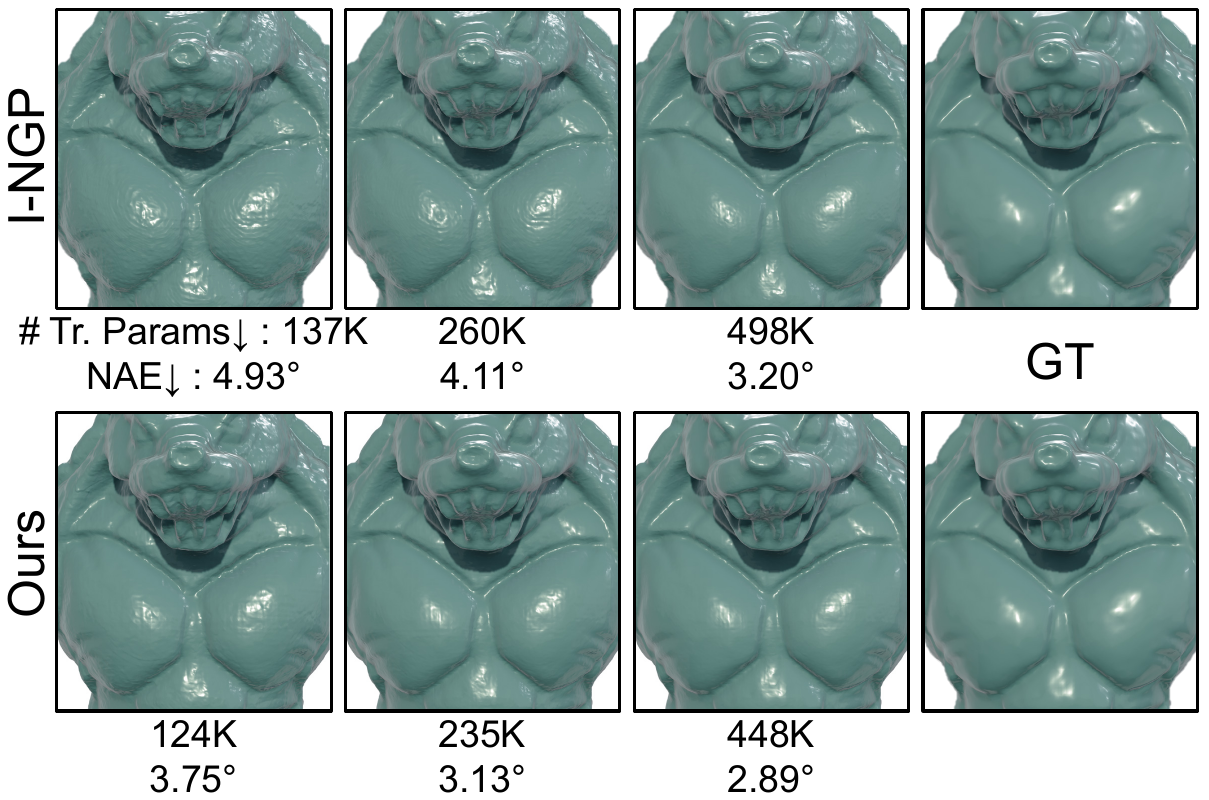}
  \caption{\textbf{3D Signed Distance Field Reconstruction.} We compare the reconstruction accuracy of Instant NGP (``I-NGP")~\cite{muller2022instant} and ours under different parameter count.}
  \label{fig:ablation_params_sdf}
\end{figure}

In Table~\ref{tab:sdf_comp}, we compare the performance under different numbers of trainable parameters. Our approach consistently has higher IoU and lower NAE. The advantages of our method are larger when using fewer parameters, which is also demonstrated in Fig.~\ref{fig:ablation_params_sdf}. 

\begin{table}[t]
\begin{center}
\resizebox*{\columnwidth}{!}{
\begin{tabular}{lcc|cc}
\toprule
 & Steps & \# Tr. Params$\downarrow$ & IoU$\uparrow$ & NAE$\downarrow$ \\
\midrule
NGLOD5~\cite{takikawa2021neural} & 245k & 10.15M & 0.9962 & 6.58 \\
NGLOD6~\cite{takikawa2021neural} & 245k & 78.84M & 0.9963 & 6.14 \\
I-NGP~\cite{muller2022instant} & 20k & 950K & 0.9994 & 5.70 \\
Ours & 20k & 856K & \textbf{0.9995} & \textbf{4.93} \\
\midrule
$\text{I-NGP}_{400K}$~\cite{muller2022instant} & 20k & 498K & 0.9992 & 6.39 \\
$\text{Ours}_{400K}$ & 20k & \textbf{448K} & 0.9994 & 5.53 \\
\bottomrule
\end{tabular}
}
\end{center}
\caption{\textbf{3D Signed Distance Field Reconstruction.} We quantitatively compare our method with NGLOD~\cite{takikawa2021neural} and Instant NGP (``I-NGP")~\cite{muller2022instant}.}
\label{tab:sdf_comp}
\end{table}

\begin{table*}[t]
\begin{center}
\resizebox{\textwidth}{!}{
\small
\begin{tabular}{lccll|llll}
\toprule
& Batch Size & Steps & Time$\downarrow$ & \# Params$\downarrow$ & PSNR$\uparrow$ & SSIM$\uparrow$ & $\text{LPIPS}_{VGG}\downarrow$ & $\text{LPIPS}_{Alex}\downarrow$ \\
\midrule
NeRF~\cite{mildenhall2021nerf} & $4096$ & $300$k & $\sim35$h & $1.25$M \tikzcircle[gold,fill=gold]{2pt} & $31.01$ & $0.947$ & $0.081$ & - \\
Mip-NeRF 360~\cite{barron2022mip} & $16384$ & $250$k & $\sim3.4$h & $3.23$M \tikzcircle[silver,fill=silver]{2pt} & $33.25$ & $0.962$ & $0.039$ \tikzcircle[silver,fill=silver]{2pt} & - \\
Point-NeRF~\cite{xu2022point} & - & $200$k & $\sim4.5$h & - & $33.31$ \tikzcircle[bronze,fill=bronze]{2pt} & $0.962$ & $0.050$ & $0.028$ \\
Plenoxels~\cite{fridovich2022plenoxels} & $5000$ & $128$k & $11.4$m \tikzcircle[silver,fill=silver]{2pt} & $194.5$M & $31.71$ & $0.958$ & $0.049$ & - \\
Instant NGP~\cite{muller2022instant} & $262144$ & $35$k & $3.8$m \tikzcircle[gold,fill=gold]{2pt} & $12.21$M & $33.18$ & $0.963$ \tikzcircle[bronze,fill=bronze]{2pt} & $0.051$ & $0.028$ \\
TensoRF~\cite{chen2022tensorf} & $4096$ & $30$k & $17.4$m & $17.95$M & $33.14$ & $0.963$ \tikzcircle[bronze,fill=bronze]{2pt} & $0.047$ \tikzcircle[bronze,fill=bronze]{2pt} & $0.027$ \tikzcircle[bronze,fill=bronze]{2pt} \\
Factor Fields~\cite{chen2023factor} & $4096$ & $30$k & $12.2$m \tikzcircle[bronze,fill=bronze]{2pt} & $5.10$M & $33.14$ & $0.961$ & - & - \\
K-Planes~\cite{fridovich2023k} & $4096$ & $30$k & $38$m & $33$M & $32.36$ & $0.962$ & $0.048$ & $0.031$ \\
\midrule
Ours & $4096$ & $30$k & $33.6$m & $17.74$M & $34.62$ \tikzcircle[gold,fill=gold]{2pt} & $0.975$ \tikzcircle[gold,fill=gold]{2pt} & $0.034$ \tikzcircle[gold,fill=gold]{2pt} & $0.018$ \tikzcircle[gold,fill=gold]{2pt} \\
$\text{Ours}_{3.66M}$ & $4096$ & $30$k & $29.3$m & $3.66$M \tikzcircle[bronze,fill=bronze]{2pt} & $33.97$ \tikzcircle[silver,fill=silver]{2pt} & $0.971$ \tikzcircle[silver,fill=silver]{2pt} & $0.039$ \tikzcircle[silver,fill=silver]{2pt} & $0.022$ \tikzcircle[silver,fill=silver]{2pt} \\
\bottomrule
\end{tabular}
}
\end{center}
\caption{\textbf{Neural Radiance Field Reconstruction.} We quantitatively compare our method with numerous state-of-the-art methods on the Synthetic NeRF dataset~\cite{mildenhall2021nerf}. Best 3 scores in each metric are marked with gold \tikzcircle[gold,fill=gold]{2pt}, silver \tikzcircle[silver,fill=silver]{2pt} and bronze \tikzcircle[bronze,fill=bronze]{2pt}.
``-" denotes the information is unavailable in the respective paper.
}
\label{tab:nerf_syn_comp}
\end{table*}

\begin{figure*}[h]
  \centering
  \includegraphics[width=1.0\textwidth]{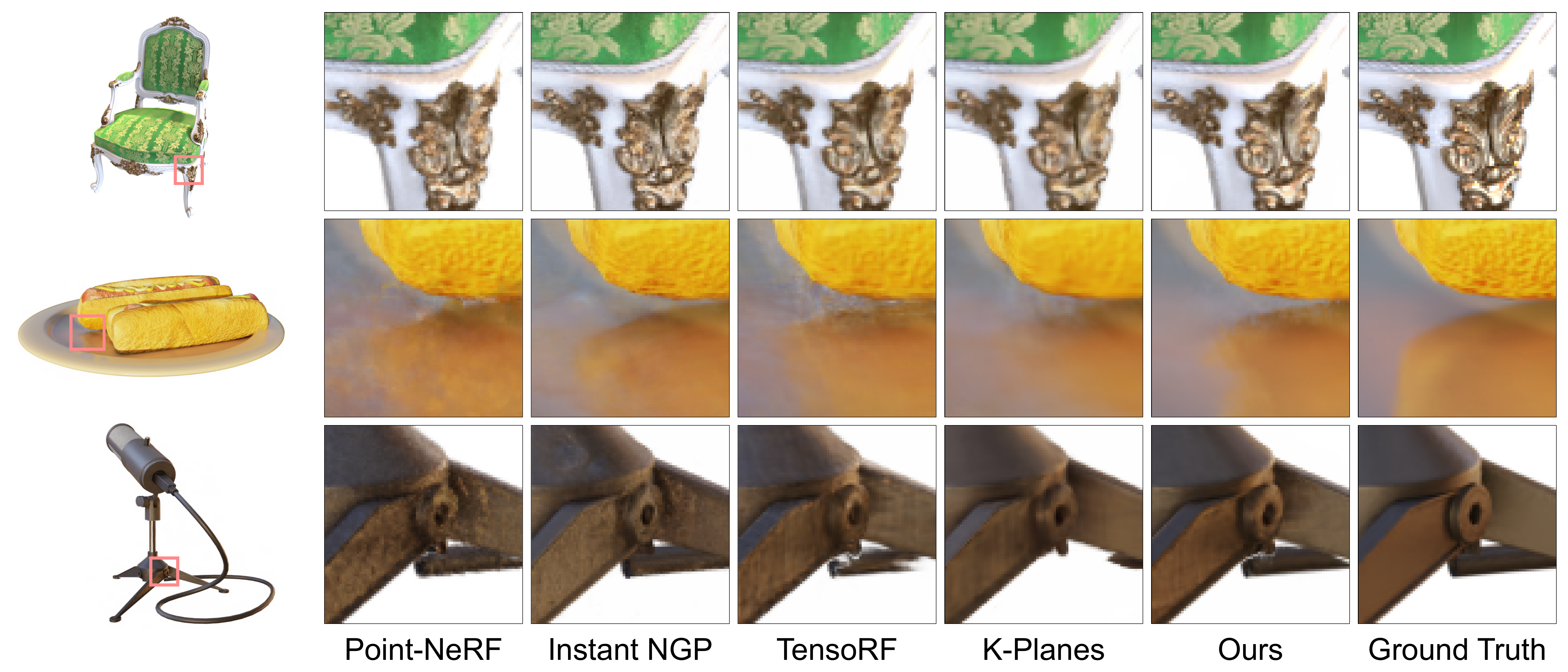}
  \caption{\textbf{Neural Radiance Field Reconstruction.}
  Qualitative comparisons on the Synthetic NeRF Dataset~\cite{mildenhall2021nerf}.
  Leftmost column shows the full-image results of our method.}
  \label{fig:nerf_syn_comp}
\end{figure*}

\begin{figure}[t]
  \centering
  \includegraphics[width=1\columnwidth]{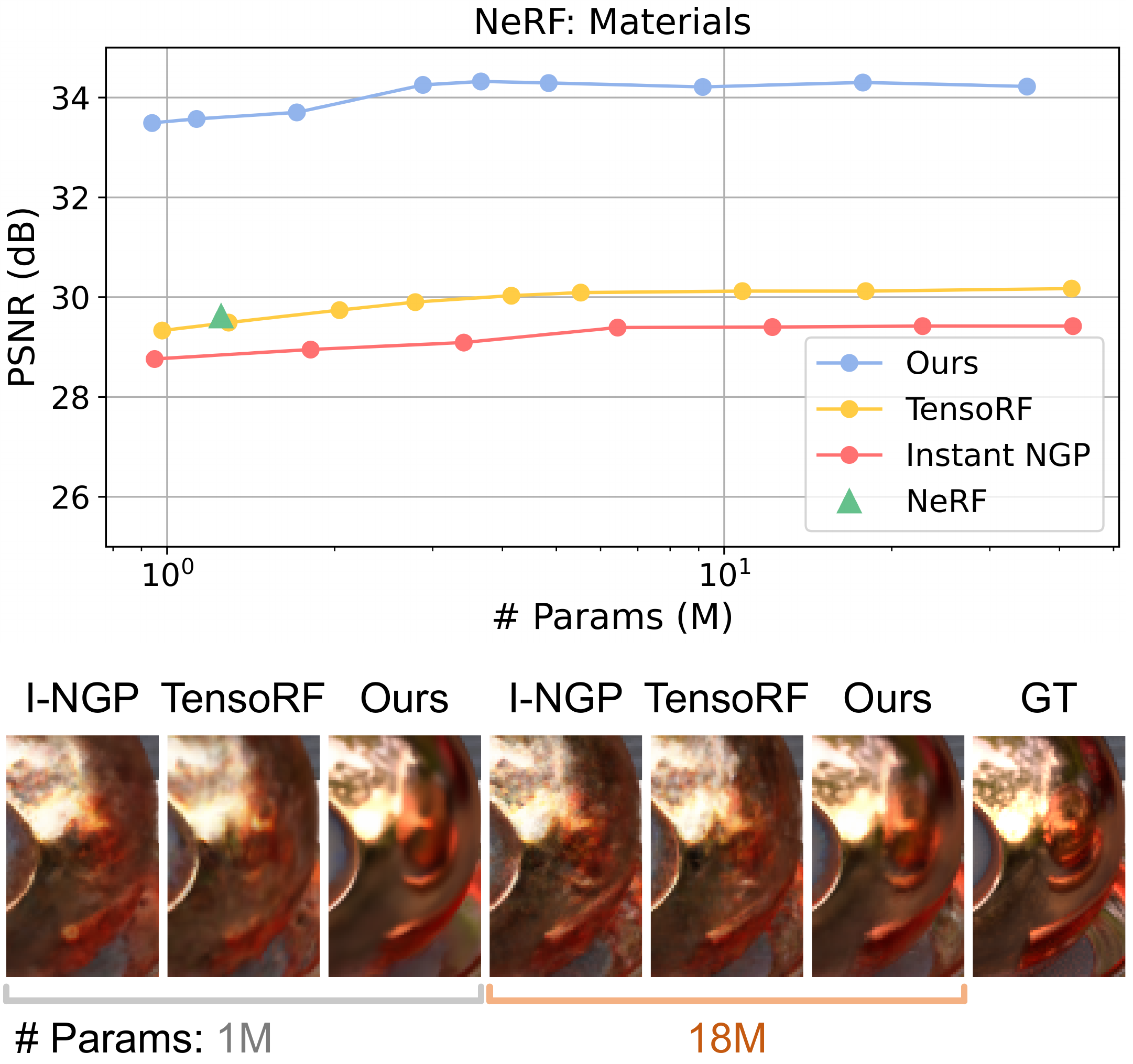}
  \caption{\textbf{Neural Radiance Field Reconstruction.} We compare the novel view synthesis quality under different parameter count on the ``Materials" scene.
  Top is a quantitative comparison of rendering PSNR. Bottom is a qualitative comparison between Instant NGP (``I-NGP")~\cite{muller2022instant}, TensoRF~\cite{chen2022tensorf} and ours at 1M and 18M parameters.}
  \label{fig:ablation_params_nerf}
\end{figure}

\subsection{Neural Radiance Field Reconstruction}
We evaluate our approach on both 360$\degree$ scenes and forward-facing scenes. Metrics of the comparison methods are taken from their paper whenever available. 
Full per-scene results are available in our supplementary material.

\paragraph{360$\degree$ Scenes.} We use the Synthetic NeRF dataset~\cite{mildenhall2021nerf} which is a widely adopted benchmark for neural radiance field reconstruction. 
We utilize TensoRF~\cite{chen2022tensorf} as the grid-based part in this experiment.
We compare with numerous representative methods in this area, as listed in Table~\ref{tab:nerf_syn_comp}.
Among them, Instant NGP~\cite{muller2022instant} and TensoRF~\cite{chen2022tensorf} represent state-of-the-art performance while Factor Fields~\cite{chen2023factor} is concurrent to our work. 
For Point-NeRF~\cite{xu2022point}, their SSIM metrics are recomputed with a consistent SSIM implementation as other work. 

Table~\ref{tab:nerf_syn_comp} comprehensively compares training time, number of parameters and novel view rendering metrics. Our method surpasses competing methods by a noticeable margin in rendering accuracy. 
Fig.~\ref{fig:nerf_syn_comp} reflects the higher quality of our results, which contain more accurate details and fewer artifacts. 
Meanwhile, our method retains a moderate model size (same as TensoRF~\cite{chen2022tensorf}) and comparable training time. After reducing to $3.66$M parameters, our model still achieves high rendering accuracy and outperforms other methods that use more parameters (Plenoxels~\cite{fridovich2022plenoxels}, Instant NGP~\cite{muller2022instant}, TensoRF~\cite{chen2022tensorf}, Factor Fields~\cite{chen2023factor}, K-Planes~\cite{fridovich2023k}). 
Fig.~\ref{fig:ablation_params_nerf} compares the novel view synthesis accuracy with representative methods (Instant NGP~\cite{muller2022instant}, TensoRF~\cite{chen2022tensorf}) under similar parameter count. 
Our method consistently performs better than the other two and also achieves higher PSNR than vanilla NeRF~\cite{mildenhall2021nerf} when using the same number of parameters.

\begin{figure}[t]
  \centering
  \includegraphics[width=1.0\columnwidth]{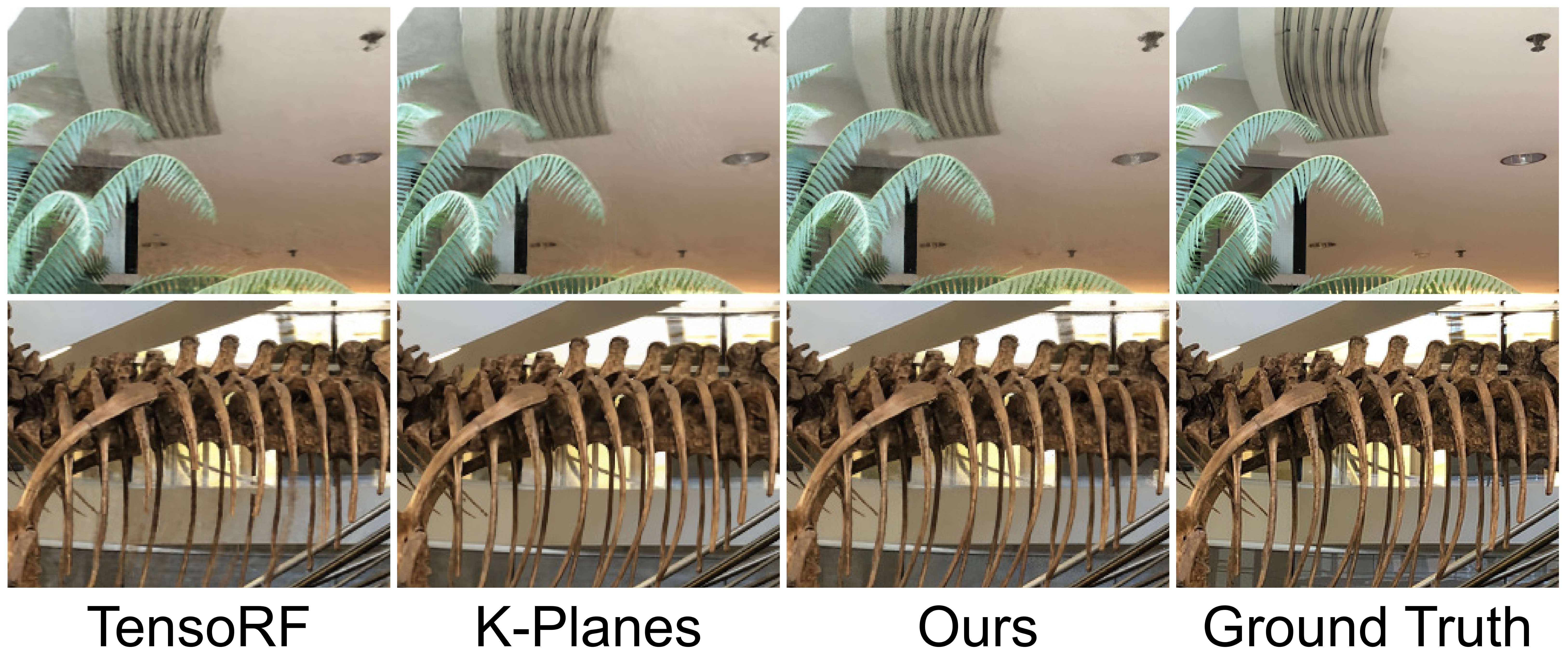}
  \caption{\textbf{Neural Radiance Field Reconstruction.} Qualitative comparisons on the LLFF Dataset~\cite{mildenhall2019local}.
  }
  \label{fig:nerf_llff_comp}
\end{figure}

\paragraph{Forward-Facing Scenes.} We use the LLFF dataset~\cite{mildenhall2019local} which contains 8 real unbounded forward-facing scenes. 
In this experiment, we explore using K-Planes~\cite{fridovich2023k} as the grid-based part .
As shown in Table~\ref{tab:nerf_llff_comp}, our approach achieves the highest PSNR and second-best SSIM. Although Mip-NeRF 360 has a higher score in SSIM, its training time is $7$ times longer than ours. Compared to Plenoxels and TensoRF, our method has higher rendering accuracy, fewer parameters and comparable training speed. Fig.~\ref{fig:nerf_llff_comp} shows example novel view synthesis results, where ours contain fewer visual artifacts. 

\begin{table}[t]
\begin{center}
\resizebox*{\columnwidth}{!}{
\begin{tabular}{lll|ll}
\toprule
& Time$\downarrow$ & \# Params$\downarrow$ & PSNR$\uparrow$ & SSIM$\uparrow$ \\
\midrule
NeRF~\cite{mildenhall2021nerf} & $36$h & $1.25$M \tikzcircle[gold,fill=gold]{2pt} & $26.50$ & $0.811$ \\
Mip-NeRF 360~\cite{barron2022mip} & $3.8$h & $3.23$M \tikzcircle[silver,fill=silver]{2pt} & $26.86$ \tikzcircle[bronze,fill=bronze]{2pt} & $0.858$ \tikzcircle[gold,fill=gold]{2pt} \\
Plenoxels~\cite{fridovich2022plenoxels} & $24$m \tikzcircle[gold,fill=gold]{2pt} & $\sim500$M & $26.29$ & $0.839$ \\
TensoRF~\cite{chen2022tensorf} & $25$m \tikzcircle[silver,fill=silver]{2pt} & $45$M & $26.73$ & $0.839$ \\
K-Planes~\cite{fridovich2023k} & $33$m & $18.7$M \tikzcircle[bronze,fill=bronze]{2pt} & $26.92$ \tikzcircle[silver,fill=silver]{2pt} & $0.847$ \tikzcircle[bronze,fill=bronze]{2pt} \\
\midrule
Ours & $31$m \tikzcircle[bronze,fill=bronze]{2pt} & $18.7$M \tikzcircle[bronze,fill=bronze]{2pt} & $27.05$ \tikzcircle[gold,fill=gold]{2pt} & $0.849$ \tikzcircle[silver,fill=silver]{2pt} \\
\bottomrule
\end{tabular}
}
\end{center}
\caption{\textbf{Neural Radiance Field Reconstruction.} Quantitative comparisons on the LLFF Dataset~\cite{mildenhall2019local}.
}
\label{tab:nerf_llff_comp}
\end{table}

\subsection{Ablation Study}

In Table~\ref{tab:ablation_arch}, we conduct ablation study on adaptive RBFs (A-RBF) and multi-frequency sinusoidal composition (MSC) using the DIV2K validation set~\cite{Agustsson_2017_CVPR_Workshops, Timofte_2017_CVPR_Workshops} and the 3D shapes in Sec.~\ref{sec:exp_sdf}. All image models are trained for $3500$ steps and all SDF models are trained for $20000$ steps.
To demonstrate the effectiveness of sinusoidal composition in our framework, we further include variants that replace it with positional encoding~\cite{mildenhall2021nerf} (Ours-PE) and sinusoid activation~\cite{sitzmann2020implicit} (Ours-SIREN). 
For Ours-PE, we apply positional encoding~\cite{mildenhall2021nerf} (PE) on input coordinate $\vb{x}$ and concatenate the features with $g_b(\vb{x})$ before input to the decoder network $g_m$. 
For Ours-SIREN, we apply sinusoidal activation~\cite{sitzmann2020implicit} to the hidden layers in $g_m$, and use the method in~\cite{sitzmann2020implicit} to initialize fully-connected layers.
As shown in Table~\ref{tab:ablation_arch}, without adaptive RBFs and sinusoidal composition, there is a noticeable drop in accuracy. Compared to PE and SIREN, our multi-frequency sinusoidal composition technique achieves higher performance.

\begin{table}[t]
\begin{center}
\resizebox*{0.95\columnwidth}{!}{
\small
\begin{tabular}{lcc|cc}
\toprule
 & \multicolumn{2}{c|}{2D Images} & \multicolumn{2}{c}{3D SDF} \\
 & PSNR$\uparrow$ & SSIM$\uparrow$ & IoU$\uparrow$ & NAE$\downarrow$ \\
\midrule
No A-RBF & 42.37 & 0.9918 & 0.9994 & 5.70 \\
No MSC on RBF & 48.19 & 0.9940 & \textbf{0.9995} & 5.04 \\
No MSC on Feat. & 48.46 & 0.9935 & \textbf{0.9995} & 5.09 \\
No MSC on Both & 43.81 & 0.9870 & \textbf{0.9995} & 5.16 \\
Ours Full & \textbf{51.53} & \textbf{0.9961} & \textbf{0.9995} & \textbf{4.93} \\
\midrule
Ours-PE & 43.72 & 0.9870 & 0.9994 & 5.46 \\
Ours-SIREN & 45.98 & 0.9920 & 0.9994 & 5.69 \\
\bottomrule
\end{tabular}
}
\end{center}
\caption{
\textbf{Ablation Study.} We ablate on the adaptive RBFs (A-RBF) and multi-frequency sinusoidal composition (MSC). ``Ours-PE" replaces MSC with positional encoding~\cite{mildenhall2021nerf}. ``Ours-SIREN" replaces MSC with sinusoid activation~\cite{sitzmann2020implicit}.
}
\label{tab:ablation_arch}
\end{table}

\section{Conclusion}
We have proposed NeuRBF, which provides accurate and compact neural representations for signals. We demonstrate that by simultaneously exploiting the spatial adaptivity and frequency extension of radial basis functions, the representation ability of neural fields can be greatly enhanced. To effectively adapt radial basis functions to target signals, we further devise tailored weighting schemes. Our method achieves higher accuracy than state-of-the-arts on 2D shape fitting, 3D signed distance field reconstruction, and neural radiance field reconstruction, while using same or fewer parameters. 
We believe our framework is a valuable step towards more expressive neural representations.

By far, we have not explored generalized learning, which would be a promising extension of our framework. Another future direction would be incorporating dictionary learning to further increase model compactness.

\section*{Acknowledgements}
The authors thank the anonymous reviewers for their valuable feedback, and Anpei Chen and Zexiang Xu for helpful discussions.

{\small
\bibliographystyle{ieee_fullname}
\bibliography{egbib}
}

\clearpage

\appendix

\section{Illustration of the Discontinuity in Adaptive RBFs}
\begin{figure}[t]
  \centering
  \includegraphics[width=1.0\columnwidth]{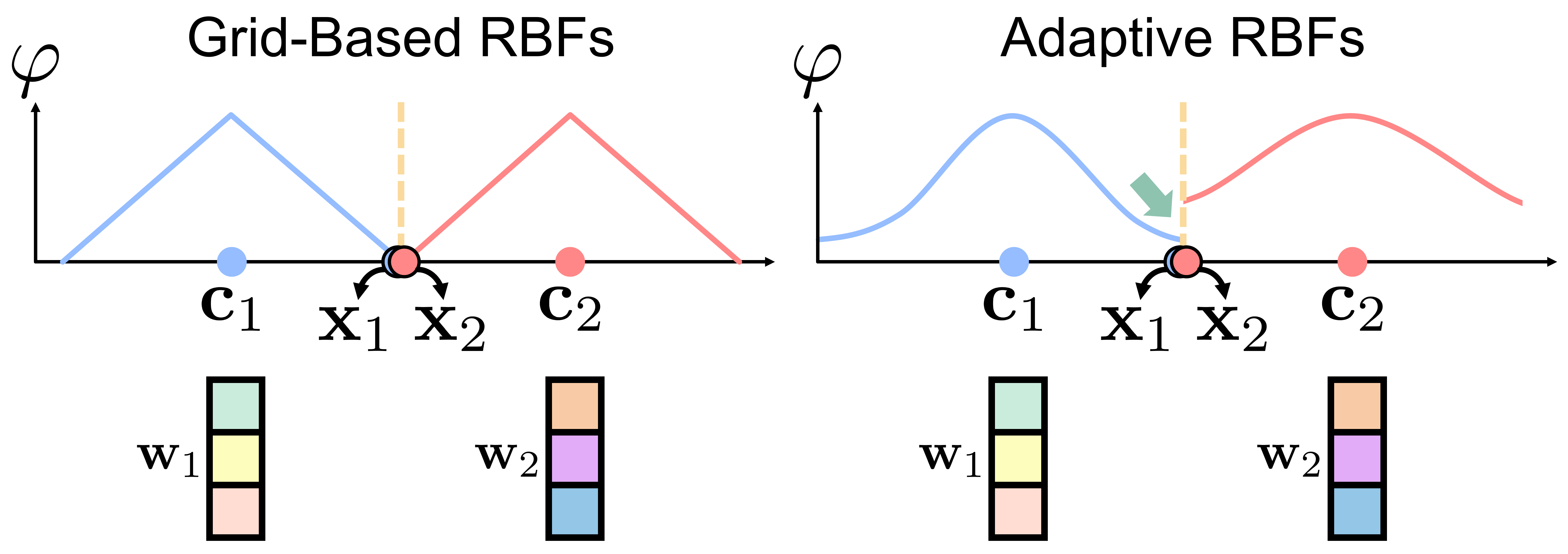}
  \caption{
  \textbf{Illustration of the discontinuity in adaptive RBFs.} 
  }
  \label{fig:discontinuity}
\end{figure}

As shown in Fig.~\ref{fig:discontinuity}, consider a simple 1D case where $\vb{x}_1 \approx \vb{x}_2$ are two points located on the boundary where $U(\vb{x})$ changes. Let $U(\vb{x}_1)=\{1\}$, $U(\vb{x}_2)=\{2\}$ be the sets of their closest RBF.
From Eq.~(2) in the paper, the aggregated neural feature $g_b(\vb{x})$ is computed as $g_b(\vb{x}) = \sum_{i \in U(\vb{x})} \varphi(\vb{x}, \vb{c}_i, \Sigma_i) \vb{w}_i$. 
Generally, for adaptive RBFs, $\varphi(\vb{x}_1, \vb{c}_1, \Sigma_1) \not\approx \varphi(\vb{x}_2, \vb{c}_2, \Sigma_2)$ and $\vb{w}_1 \not\approx \vb{w}_2$. Therefore, $g_b(\vb{x}_1) \not\approx g_b(\vb{x}_2)$. This reveals a discontinuity in $g_b(\vb{x})$ when $\vb{x}$ changes from $\vb{x}_1$ to $\vb{x}_2$. On the other hand, for grid-based RBFs that use linear interpolation as kernel function, both $\varphi(\vb{x}_1, \vb{c}_1)$ and $\varphi(\vb{x}_2, \vb{c}_2)$ are close to $0$, so $g_b(\vb{x})$ does not contain such discontinuity. 
We combine adaptive and grid-based RBFs through feature concatenation to balance fitting accuracy and interpolation smoothness.

\section{Details on RBF Initialization}
We utilize the EM-style Lloyd’s K-Means algorithm to initialize RBF positions using all points. The number of RBFs is calculated based on parameter budget. The initialization is conducted only once per scene, before the start of training. We do not split or merge RBFs during training.
During weighted K-Means, the initial centers are generated by weighted random sampling. 
We do not repeat this random sampling for multiple times because we observe it does not have major influence on final performance.
The E-M steps are the following:
\begin{equation}
a_{ij} = 
\begin{cases}
    1, \text{if } i = \underset{k}{\arg\min} \lVert \vb{x}_j - \vb{c}_k \rVert^2, \\
    0,              \text{otherwise}. \\
\end{cases}\\
\label{eq:kmeans_em1}
\end{equation}
\begin{equation}
\vb{c}_i = \frac{\sum_j a_{ij} w_j \vb{x}_j}{\sum_j a_{ij} w_j}.
\label{eq:kmeans_em2}
\end{equation}
$a_{ij}$ is an indicator variable: $a_{ij}=1$ if $\vb{x}_j$ is assigned to cluster $i$ and $a_{ij}=0$ otherwise. 
For efficiency, we iterate Eq.~\eqref{eq:kmeans_em1}\eqref{eq:kmeans_em2} for only $10$ steps as the results are already close to convergence and sufficient for our use case. 
We implement the E-M steps with parallel KD Tree and vectorized centroid update.

\section{More Ablation Study}
\subsection{RBF Initialization}
\begin{figure*}[t]
  \centering
  \includegraphics[width=0.9\textwidth]{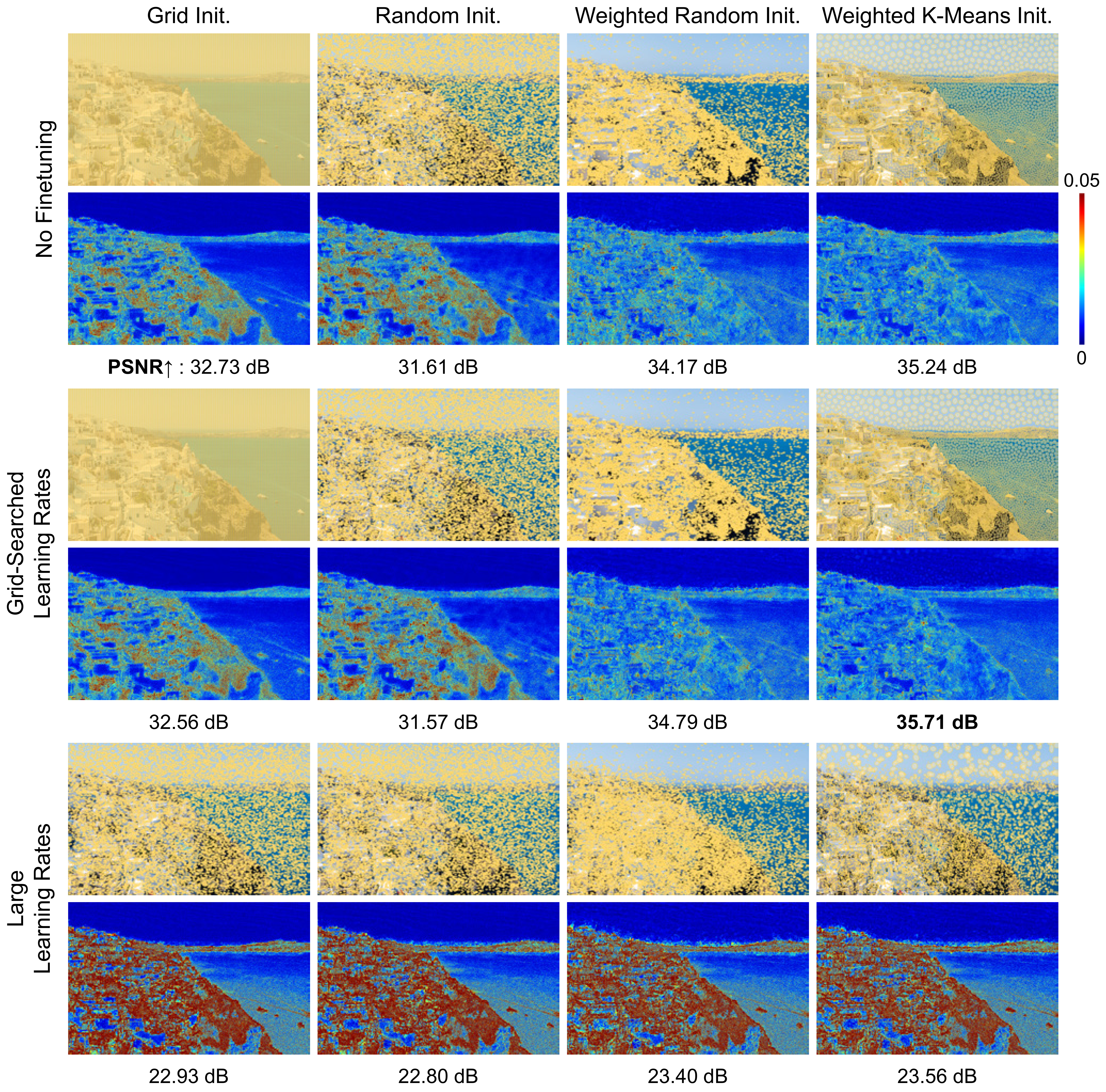}
  \caption{
  \textbf{Evaluation on RBF Initialization.} 
  We compare different RBF initialization methods in columns: grid initialization, random initialization, weighted random initialization and weighted K-Means initialization. We also evaluate different RBF finetuning strategies using gradient backpropagation (with only L2 loss on pixel value) in rows: no finetuning, grid-searched learning rates, large learning rates. For each result, we visualize the RBF parameters as yellow ellipses and show fitting error maps.
  }
  \label{fig:ablation_init_img}
\end{figure*}

To evaluate the effects of RBF initialization, we compare weighted K-Means with grid initialization, random initialization and weighted random initialization. As shown in Fig.~\ref{fig:ablation_init_img}, we use an image from DIV2K dataset~\cite{Agustsson_2017_CVPR_Workshops, Timofte_2017_CVPR_Workshops} and conduct 2D image fitting. 
To facilitate visualization, we only use $15129$ RBFs in each baseline.
We visualize the position and shape parameters of RBFs as yellow ellipses, and show the fitting error maps and PSNR. 
As demonstrated in the top two rows, weighted K-Means initialization achieves the highest fitting accuracy. Among the other three baselines, weighted random initialization has a competitve performance while random initialization leads to the worst result.

In the bottom four rows, we further evaluate the effectiveness of using gradient backpropagation to finetune RBF parameters during training. We first use a set of reasonable learning rates for position and shape parameters, which are obtained through grid search on the baseline with weighted random initialization. 
As shown in the middle two rows, gradient backpropagation (with only L2 loss on pixel value) only provides minor improvement compared to the first two rows. Besides, the update to the RBF parameters is barely noticeable for grid initialization.
Then, we experiment with large learning rates in the last two rows. It can be seen that the RBF parameters can be largely changed from their initialization. However, this leads to significant performance drop for all baselines.
The above results validate the benefits of RBF initialization.

\begin{figure*}[t]
  \centering
  \includegraphics[width=0.9\textwidth]{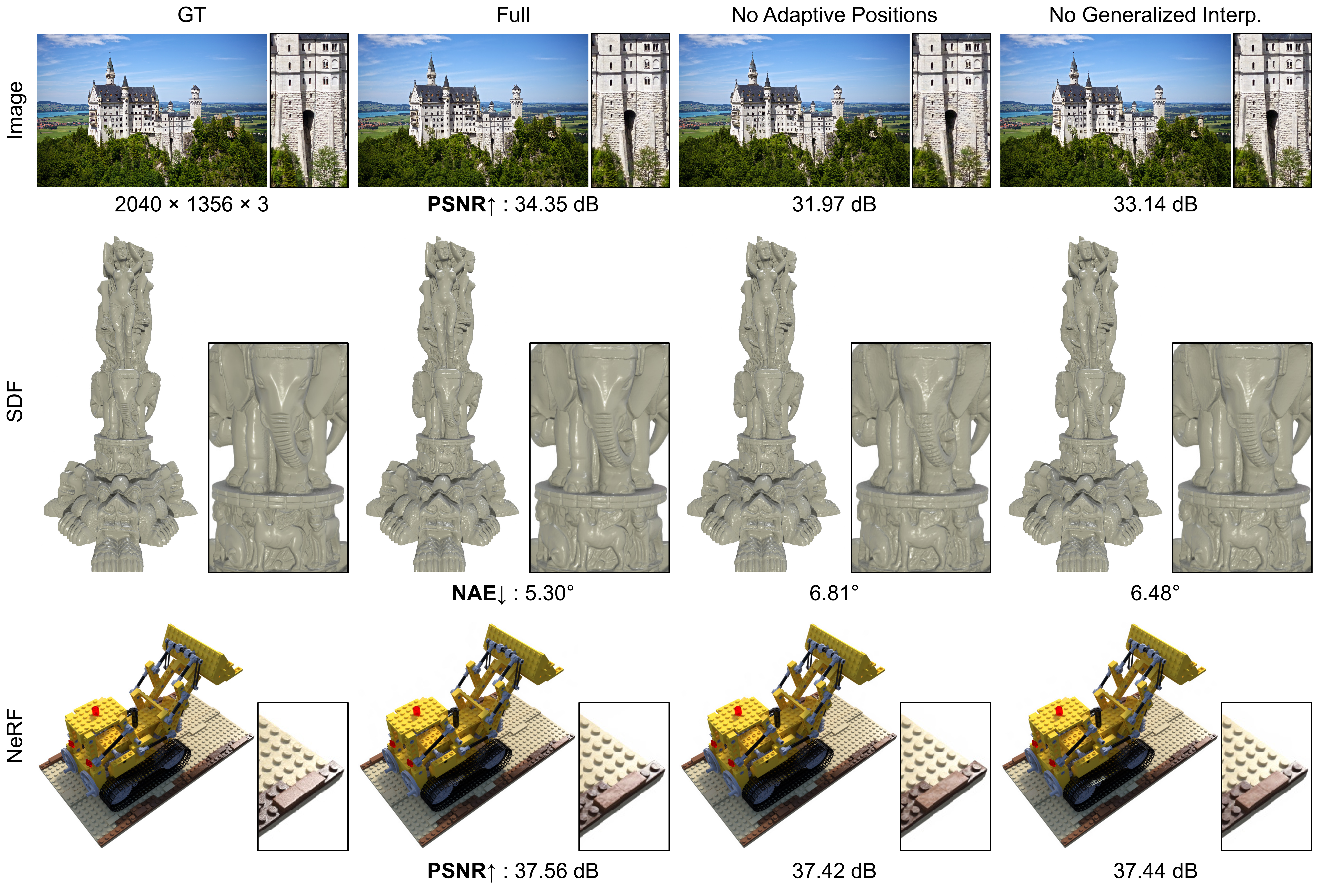}
  \caption{
  \textbf{Evaluation on the adaptive positions and generalized interpolation of RBFs.} 
  ``No Adaptive Positions": the positions of RBFs are fixed to a grid structure. 
  ``No Generalized Interp.": the interpolation function is N-dimensional linear interpolation. 
  }
  \label{fig:ablation_kcs}
\end{figure*}

\subsection{Adaptive Positions and Generalized Interpolation}
Here, we evaluate the effects of using adaptive positions for RBFs and generalizing N-dimensional linear interpolation to RBFs with shape parameters. We conduct this ablation study on image, SDF and NeRF tasks, and the results are shown in Fig.~\ref{fig:ablation_kcs}. 
The parameter count of each model is 567K, 856K and 17.7M respectively for the three tasks.
Based on the results, both adaptive positions and generalized interpolation are beneficial to performance.

\section{More Implementation Details}
\paragraph{Architecture.}
For the decoder network $g_m$, except NeRF task, we use a 3-layer MLP (2 hidden layers + 1 output layer) with a network width of 64 neurons, where rectified linear unit (ReLU) activation function is applied to the second hidden layer. 
The MLP uses a very small part of the parameters (\eg, only 7K in image fitting). 
For the NeRF experiments on the Synthetic NeRF dataset~\cite{mildenhall2021nerf}, we use a single Softplus layer as density decoder (same as TensoRF~\cite{chen2022tensorf}) and use the rendering equation encoding from NRFF~\cite{han2023multiscale} as color decoder. 
For the NeRF experiments on the real LLFF Forward-Facing dataset~\cite{mildenhall2019local}, we adopt the same network architecture as K-Planes-hybrid~\cite{fridovich2023k}, which uses a 2-layer MLP for density decoder and a 3-layer MLP for color decoder.

\paragraph{Experiments on 2D Image Fitting.}
The neural features $\vb{w}_i$ of adaptive RBFs have a channel dimension of 32.
The neighboring RBFs $U(\vb{x})$ of a point $\vb{x}$ is its $4$ nearest neighbors.
For Instant NGP~\cite{muller2022instant}, we use their official open-sourced codes and hyper-parameters in the comparison experiments. 
Note that some results in their paper use smaller hash table sizes, hence fewer parameters but also lower PSNR.
For MINER~\cite{saragadam2022miner}, we use the implementation from MINER\_pl\footnote[1]{\url{https://github.com/kwea123/MINER_pl}}. The original MINER paper does not report their results on the DIV2K dataset~\cite{Agustsson_2017_CVPR_Workshops, Timofte_2017_CVPR_Workshops}.
All methods use a batch size of $262144$ and are trained for a same number of steps. 

\paragraph{Experiments on 3D SDF Reconstruction.}
The neural features $\vb{w}_i$ of adaptive RBFs have a channel dimension of 16, and the size of neighborhood $U(\vb{x})$ is 8.
We use a grid resolution of $1024^3$ for marching cubes. IoU is evaluated on these grid points.
For normal angular error (NAE), it is computed similarly as normal consistency~\cite{chibane2020implicit}, but is in unit of degree. Specifically, let $P_1, P_2$ be randomly sampled points on two mesh surfaces, NN$(x, P)$ be the closest point to $x$ in point set $P$, and $n_f(x)\in\mathbb{R}^{3\times1}$ be the unit face normal of point $x$, NAE is calculated as:
\begin{equation}
\begin{split}
& \text{NAE}(P_1, P_2) = \frac{1}{2} \cdot \frac{180}{\pi} \cdot ( \\
& \frac{1}{|P_1|}\sum_{p_1 \in P_1}\arccos(n_f(p_1)^T n_f(\text{NN}(p_1, P_2))) + \\
& \frac{1}{|P_2|}\sum_{p_2 \in P_2}\arccos(n_f(p_2)^T n_f(\text{NN}(p_2, P_1)))).
\label{eq:nae}
\end{split}
\end{equation}

\paragraph{Experiments on Neural Radiance Field Reconstruction.}
For the Synthetic NeRF dataset~\cite{mildenhall2021nerf}, the channel dimension of adaptive RBFs is 32 and the size of neighborhood $U(\vb{x})$ is 5.
We first train the grid-based part for $1000$ steps, which is then used to distill scene information and conduct RBF initialization. 
For the real LLFF Forward-Facing dataset~\cite{mildenhall2019local}, the channel dimension of adaptive RBFs is 16 and the size of neighborhood $U(\vb{x})$ is 8.
The grid-based model used for distillation is trained for $2000$ steps while the full model is trained for $38000$ steps.

\begin{table*}[t]
\begin{center}
\resizebox{\textwidth}{!}{
\begin{tabular}{l|c|cccccccccc}
\toprule
 & Avg. & Armadillo & Bunny & Dragon & Buddha & Lucy & XYZ Dragon & Statuette & David & Chameleon & Mechanism \\
\midrule
\midrule
\multicolumn{2}{l}{\textbf{NAE$\downarrow$}} \\
\midrule
NGLOD5~\cite{takikawa2021neural} & 6.58 & 3.60 & 4.81 & 2.85 & 3.28 & 4.73 & 5.66 & 7.53 & 3.43 & 15.91 & 14.00 \\
NGLOD6~\cite{takikawa2021neural} & 6.14 & 3.35 & 4.47 & 2.76 & 3.02 & 4.28 & 5.15 & 6.46 & 3.22 & 14.64 & 14.03 \\
I-NGP~\cite{muller2022instant} & 5.70 & 2.89 & \textbf{1.96} & 2.30 & 2.73 & 3.57 & 4.51 & 6.00 & 2.88 & 11.96 & 18.21 \\
Ours & \textbf{4.93} & \textbf{2.83} & 2.00 & \textbf{2.22} & \textbf{2.69} & \textbf{3.36} & \textbf{4.14} & \textbf{5.30} & \textbf{2.62} & \textbf{10.42} & \textbf{13.73} \\
\midrule
$\text{I-NGP}_{400K}$~\cite{muller2022instant} & 6.39 & 3.20 & 2.22 & 2.64 & 3.18 & 4.29 & 4.96 & 6.82 & 3.27 & 13.04 & 20.31 \\
$\text{Ours}_{400K}$ & 5.53 & 2.89 & 2.14 & 2.35 & 2.88 & 3.70 & 4.44 & 6.07 & 2.85 & 11.96 & 16.00 \\
\midrule
\midrule
\multicolumn{2}{l}{\textbf{IoU$\uparrow$}} \\
\midrule
NGLOD5~\cite{takikawa2021neural} & 0.9962 & 0.99974 & 0.97664 & 0.99964 & 0.99977 & 0.99979 & 0.99981 & 0.99969 & 0.99960 & 0.99456 & 0.99237 \\
NGLOD6~\cite{takikawa2021neural} & 0.9963 & 0.99979 & 0.97696 & 0.99969 & 0.99977 & 0.99986 & 0.99983 & 0.99980 & 0.99963 & 0.99528 & 0.99237 \\
I-NGP~\cite{muller2022instant} & 0.9994 & \textbf{0.99997} & 0.99968 & \textbf{0.99995} & \textbf{0.99996} & \textbf{0.99997} & \textbf{0.99996} & 0.99993 & \textbf{0.99993} & \textbf{0.99893} & 0.99605 \\
Ours & \textbf{0.9995} & 0.99994 & 0.99943 & \textbf{0.99995} & \textbf{0.99996} & 0.99996 & \textbf{0.99996} & \textbf{0.99995} & \textbf{0.99993} & 0.99765 & \textbf{0.99837} \\
\midrule
$\text{I-NGP}_{400K}$~\cite{muller2022instant} & 0.9992 & 0.99995 & \textbf{0.99974} & 0.99994 & 0.99994 & 0.99996 & 0.99995 & 0.99990 & 0.99990 & 0.99820 & 0.99448 \\
$\text{Ours}_{400K}$ & 0.9994 & 0.99996 & 0.99964 & \textbf{0.99995} & 0.99995 & \textbf{0.99997} & 0.99995 & 0.99991 & 0.99992 & 0.99706 & 0.99767 \\
\bottomrule
\end{tabular}
}
\end{center}
\caption{\textbf{3D Signed Distance Field Reconstruction.} Per-object breakdown of the quantitative metrics (NAE$\downarrow$ and IoU$\uparrow$) in Table 2 of the paper.
}
\label{tab:sdf_comp_full}
\end{table*}

\section{Limitations and Future Work}
In this work, we have primarily focused on local neural representation. It could be promising to explore the combination with other activation functions in MLP (\eg, WIRE~\cite{saragadam2023wire}). 
Besides, in our current implementation, the multipliers $\vb{m}, \vb{m}_0$ are treated as hyper-parameters and are not trainable. We tried training them along with other parameters, but observed little improvement. A possible reason is that they act as frequencies and would require tailored optimization techniques.

Our method demonstrates high representation accuracy in spatial domains; however, similar to Instant NGP~\cite{muller2022instant} and Factor Fields~\cite{chen2023factor}, we have not explored spatial-temporal tasks such as dynamic novel view synthesis. By extending radial basis functions into higher dimensions or using dimension decomposition techniques, our method can potentially be applied to these tasks. 
We also observe that it is difficult to represent large-scale complicated signals with both high accuracy and small model size, which is a common challenge for local neural fields methods.
An interesting future direction would be to design basis functions with more adaptive shapes and long-range support.

\section{Additional Results}
\subsection{2D Image Fitting}
\begin{figure*}[t]
  \centering
  \includegraphics[width=0.86\textwidth]{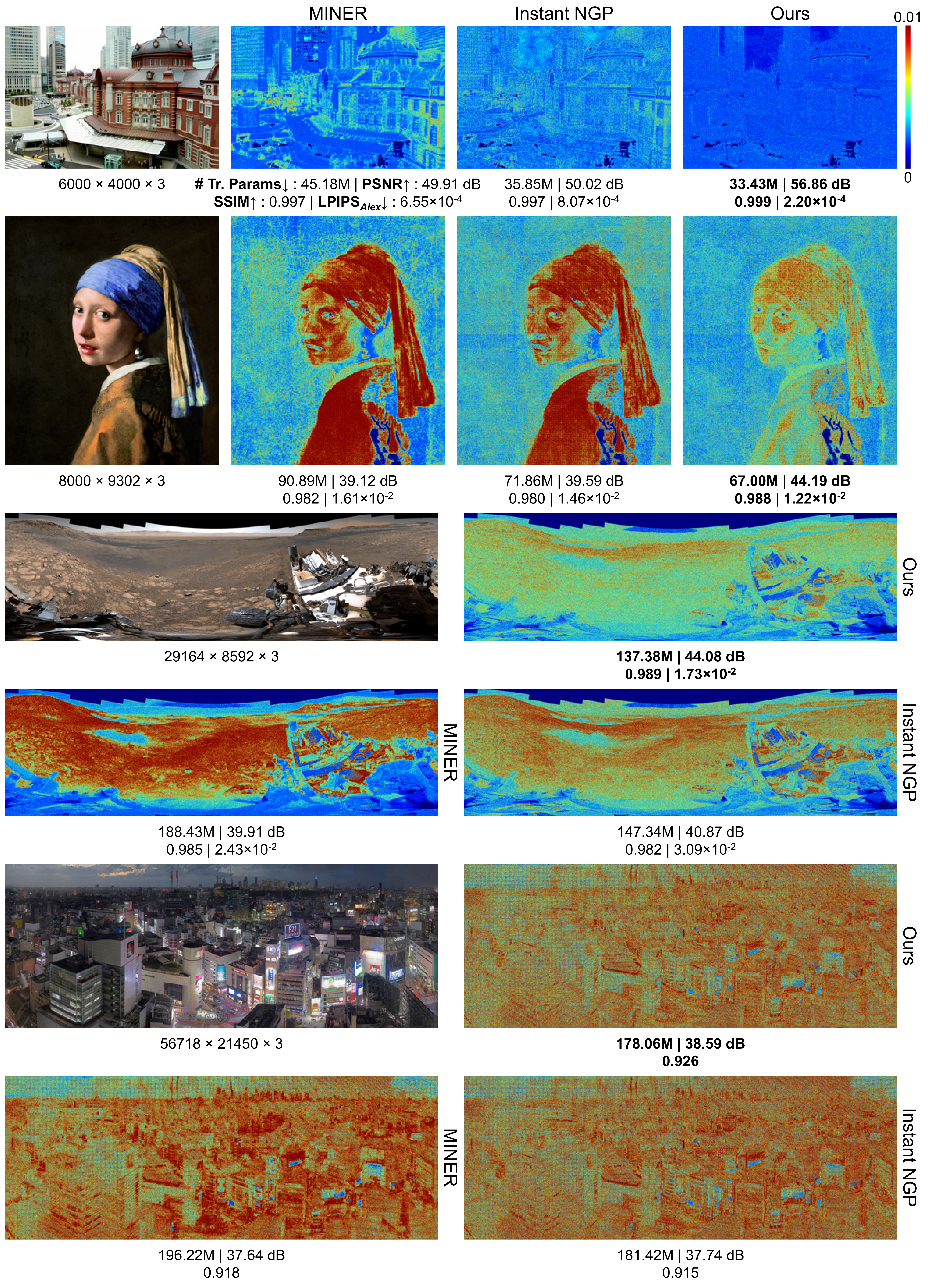}
  \caption{\textbf{2D Image Fitting.} Leftmost column or top left quarter shows the fitted images of our method and the resolution of the images. The other columns or quarters show the error maps of each method, along with the number of trainable parameters (``\# Tr. Params")$\downarrow$, PSNR$\uparrow$, SSIM$\uparrow$ and LPIPS$_{Alex}$$\downarrow$. For the last image, its resolution is too high to compute LPIPS$_{Alex}$.
  ``Girl With a Pearl Earring" renovation $\copyright$Koorosh Orooj (CC BY-SA 4.0).
  }
  \label{fig:img_comp_full}
\end{figure*}

Fig.~\ref{fig:img_comp_full} compares the results on 4 ultra-high resolution images that are not displayed in the paper due to page limit. For the error maps, we calculate the mean absolute error across color channels for each pixel. To highlight the difference among methods, we set the color bar range as $0 \sim 0.01$ (the range of pixel value is $0 \sim 1$). 

For the Pluto image (Fig. 4 row 2 in the paper), when fitting the 16 megapixel version of it, our method can reach $44.13$ dB PSNR with 7.8M parameters and 50s training.

\subsection{3D SDF Reconstruction}
\begin{figure*}[t]
  \centering
  \includegraphics[width=1\textwidth]{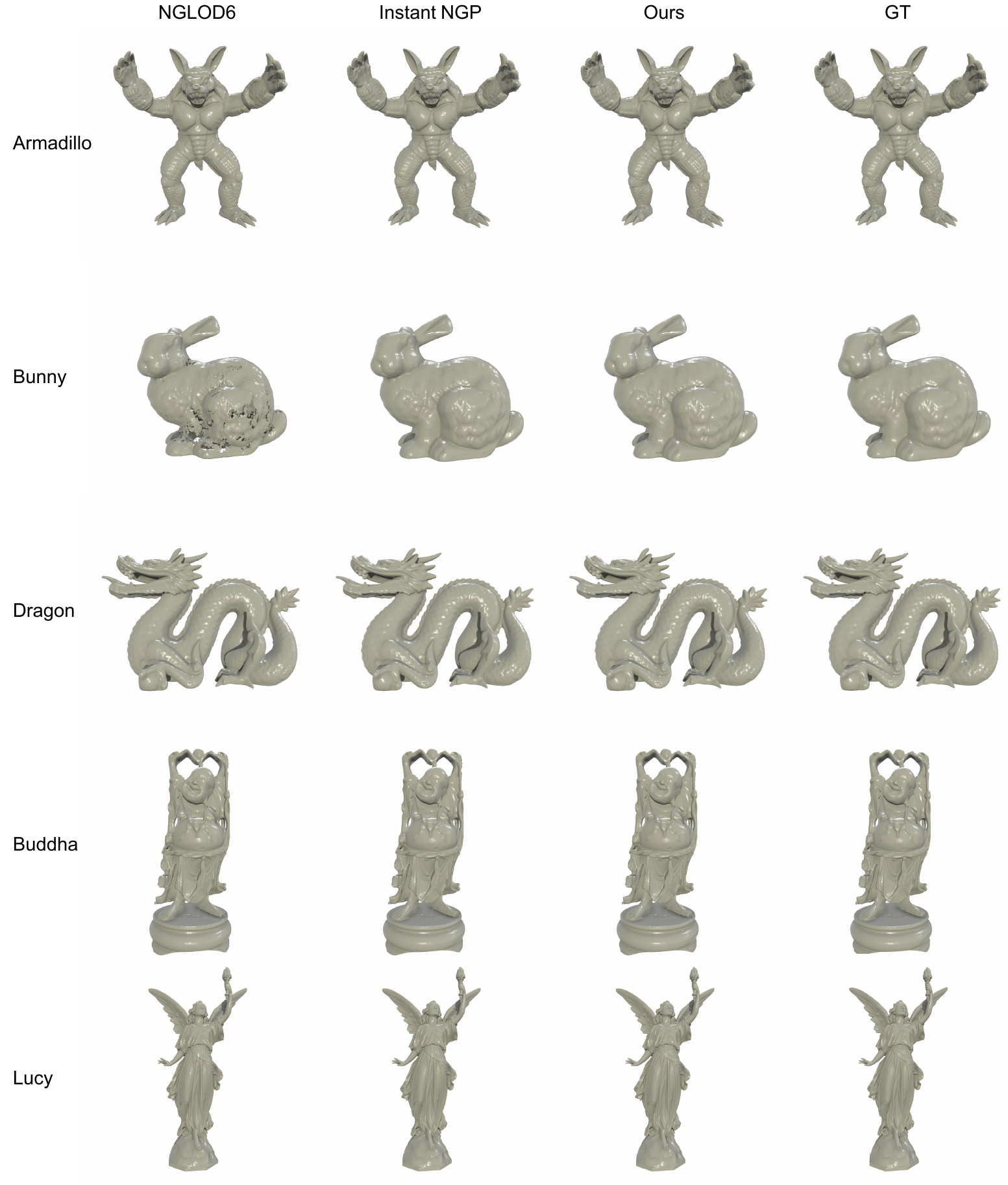}
  \caption{\textbf{3D SDF Reconstruction.} Qualitative comparisons between NGLOD6~\cite{takikawa2021neural}, Instant NGP~\cite{muller2022instant} and ours. 
  For the results in this figure, the number of trainable parameters of Instant NGP is 950K, while that of ours is 856K. (To be continued in the next page.)
  }
  \label{fig:sdf_comp_full1}
\end{figure*}

\begin{figure*}[t]
  \centering
  \includegraphics[width=1\textwidth]{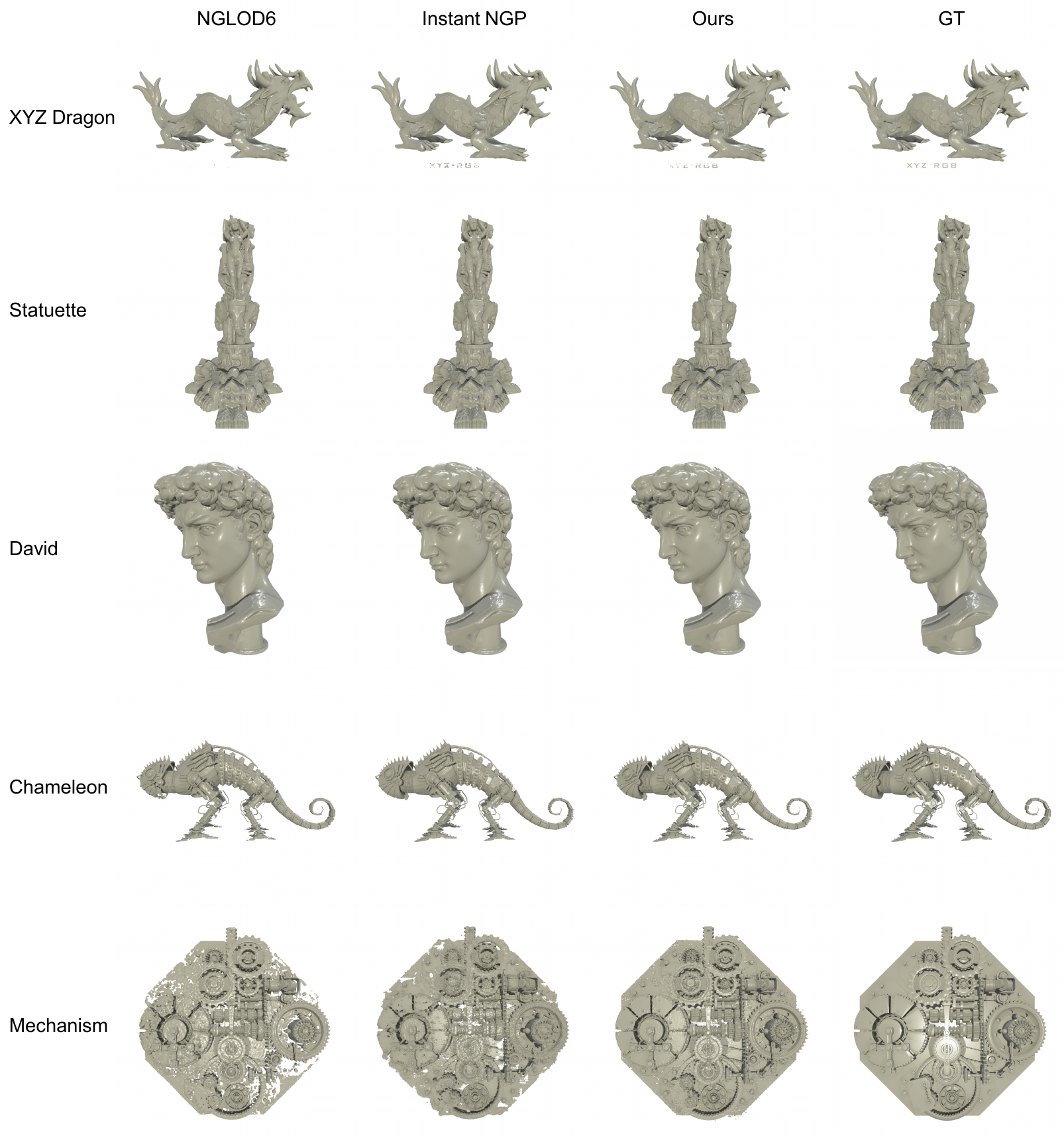}
  \caption{\textbf{3D SDF Reconstruction.} Qualitative comparisons between NGLOD6~\cite{takikawa2021neural}, Instant NGP~\cite{muller2022instant} and ours. 
  For the results in this figure, the number of trainable parameters of Instant NGP is 950K, while that of ours is 856K.
  }
  \label{fig:sdf_comp_full2}
\end{figure*}

Table~\ref{tab:sdf_comp_full} shows per-object breakdown of the quantitative metrics (NAE$\downarrow$ and IoU$\uparrow$) in Table 2 of the paper. Fig.~\ref{fig:sdf_comp_full1},~\ref{fig:sdf_comp_full2} show the qualitative results, where the numbers of trainable parameters for Instant NGP and ours are 950K and 856K.

We further compare with BACON~\cite{lindell2022bacon} and let our method use the same training settings as them.
BACON uses 531K parameters while our models only use 448K.
Averaging over 4 scenes (Armadillo, Lucy, XYZ Dragon, Statuette), the normal angular errors (NAE$\downarrow$) are 5.89\degree (BACON) vs. \textbf{4.53\degree (Ours)}. 

\subsection{Neural Radiance Field Reconstruction}
\begin{figure*}[t]
  \centering
  \includegraphics[width=1.0\textwidth]{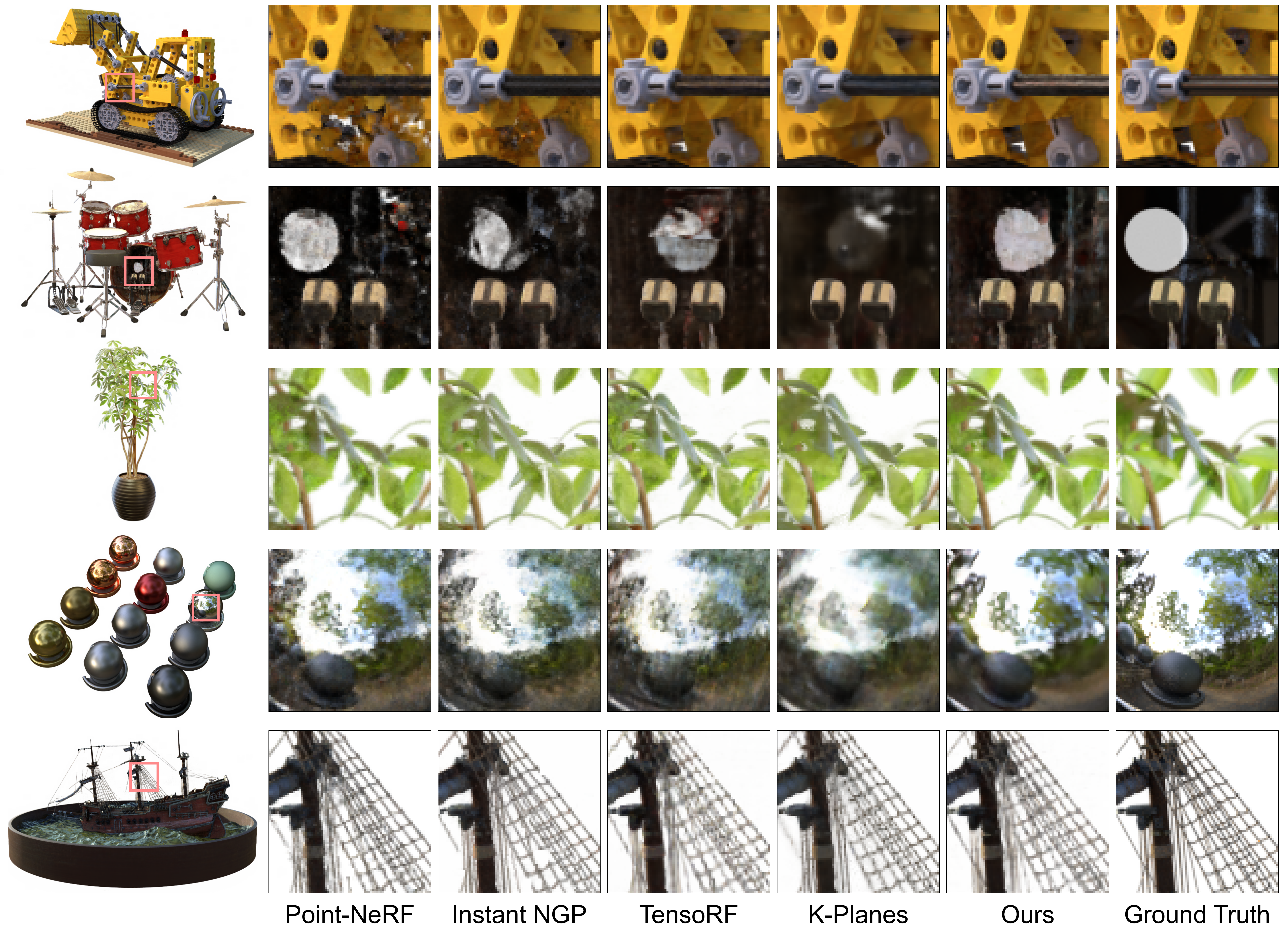}
  \caption{\textbf{Neural Radiance Field Reconstruction.}
  More close-up comparisons on the Synthetic NeRF Dataset~\cite{mildenhall2021nerf}.
  Leftmost column shows the full-image results of our method.}
  \label{fig:nerf_syn_comp_more}
\end{figure*}

\begin{figure*}[t]
  \centering
  \includegraphics[width=1.0\textwidth]{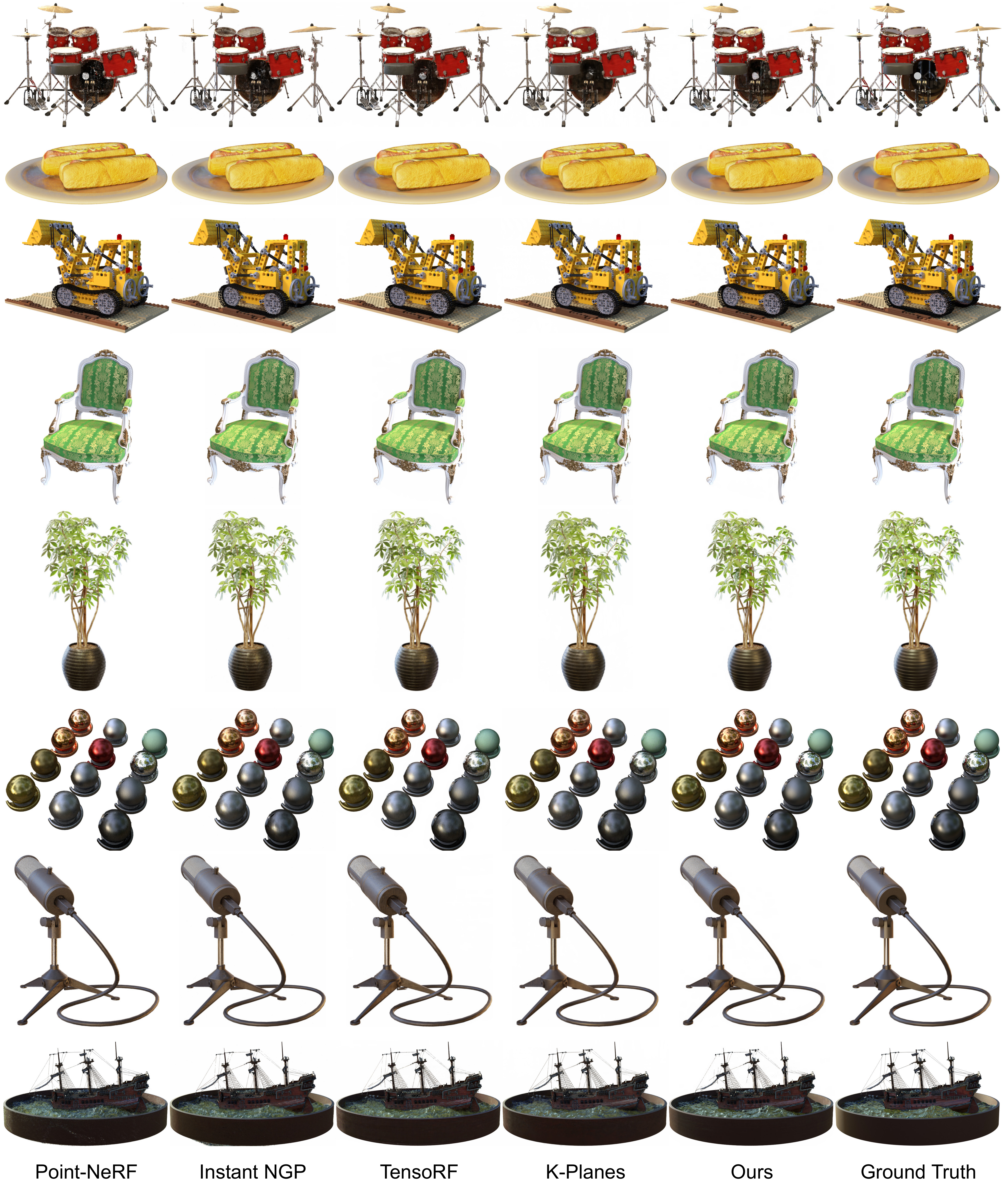}
  \caption{\textbf{Neural Radiance Field Reconstruction.}
  Full-image comparisons on the Synthetic NeRF Dataset~\cite{mildenhall2021nerf}.
  }
  \label{fig:nerf_syn_comp_full}
\end{figure*}

\begin{figure*}[t]
  \centering
  \includegraphics[width=0.81\textwidth]{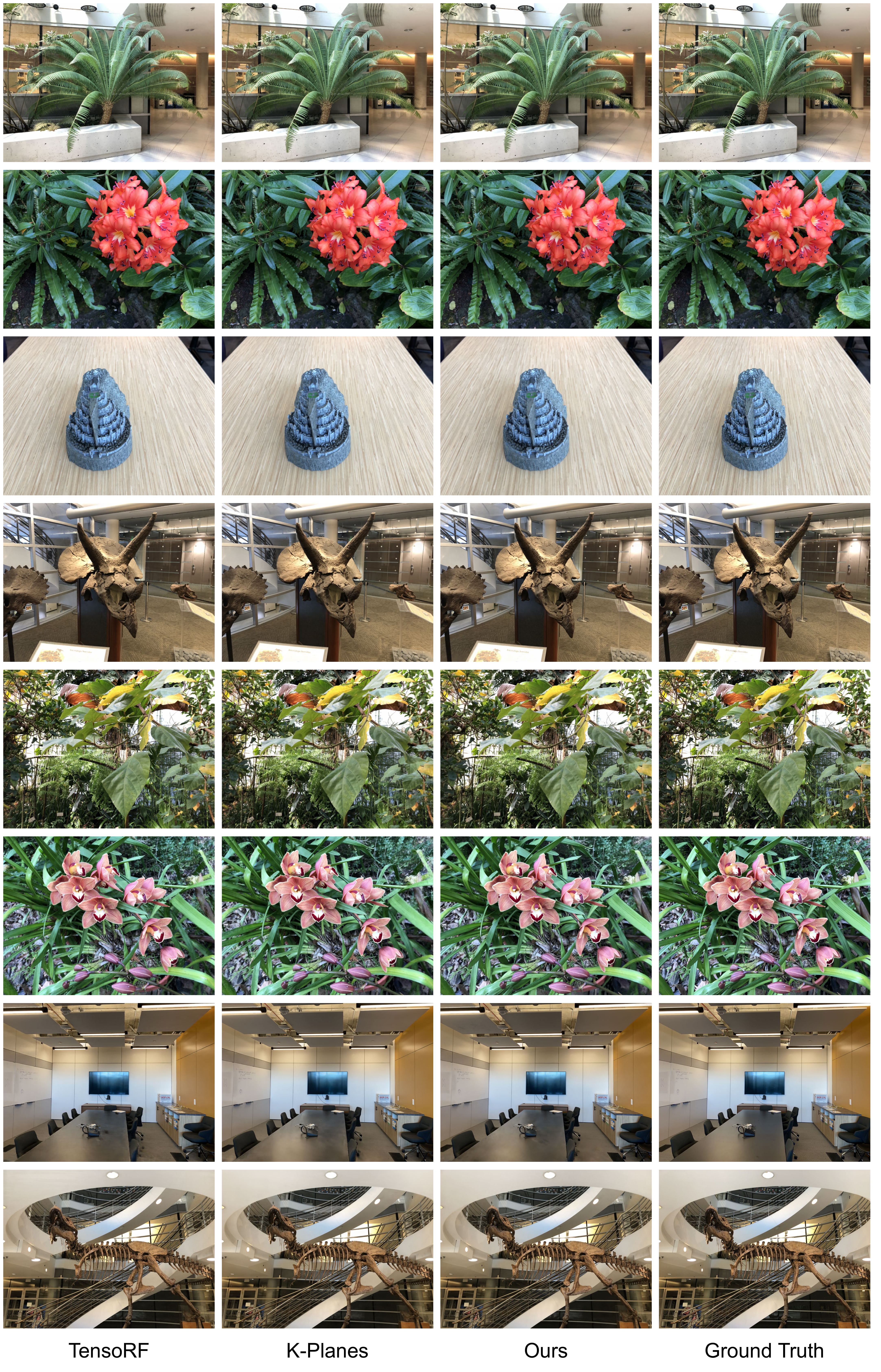}
  \caption{
  \textbf{Neural Radiance Field Reconstruction.} 
  Full-image comparisons on the real LLFF Forward-Facing dataset~\cite{mildenhall2019local}.
  }
  \label{fig:nerf_llff_comp_full}
\end{figure*}

Table~\ref{tab:nerf_syn_perscene} and~\ref{tab:nerf_llff_perscene} demonstrate the per-scene quantitative comparisons (PSNR$\uparrow$, SSIM$\uparrow$, LPIPS$_{VGG}$$\downarrow$, LPIPS$_{Alex}$$\downarrow$) on the Synthetic NeRF dataset~\cite{mildenhall2021nerf} and the real LLFF Forward-Facing dataset~\cite{mildenhall2019local}. Fig.~\ref{fig:nerf_syn_comp_more} and Fig.~\ref{fig:nerf_syn_comp_full} show more close-up and full-image comparisons on the Synthetic NeRF dataset~\cite{mildenhall2021nerf}.
Fig.~\ref{fig:nerf_llff_comp_full} shows full-image comparisons on the real LLFF Forward-Facing dataset~\cite{mildenhall2019local}.

\clearpage
\clearpage

\clearpage
\clearpage

\clearpage
\clearpage

\clearpage
\begin{table*}[t]
\begin{center}
\resizebox{\textwidth}{!}{
\begin{tabular}{l|l|llllllll}
\toprule
Methods & Avg. & Chair & Drums & Ficus & Hotdog & Lego & Materials & Mic & Ship \\
\midrule
\midrule
\multicolumn{2}{l}{\textbf{PSNR$\uparrow$}} \\
\midrule
NeRF~\cite{mildenhall2021nerf} & $31.01$ & $33.00$ & $25.01$ & $30.13$ & $36.18$ & $32.54$ & $29.62$ & $32.91$ & $28.65$ \\
Mip-NeRF 360~\cite{barron2022mip} & $33.25$ & - & - & - & - & - & - & - & - \\
Point-NeRF~\cite{xu2022point} & $33.31$ \tikzcircle[bronze,fill=bronze]{2pt} & $35.40$ & $26.06$ \tikzcircle[bronze,fill=bronze]{2pt} & $36.13$ \tikzcircle[gold,fill=gold]{2pt} & $37.30$ & $35.04$ & $29.61$ & $35.95$ \tikzcircle[bronze,fill=bronze]{2pt} & $30.97$ \\
Plenoxels~\cite{fridovich2022plenoxels} & $31.71$ & $33.98$ & $25.35$ & $31.83$ & $36.43$ & $34.10$ & $29.14$ & $33.26$ & $29.62$ \\
Instant NGP~\cite{muller2022instant} & $33.18$ & $35.00$ & $26.02$ & $33.51$ & $37.40$ & $36.39$ & $29.78$ & $36.22$ \tikzcircle[gold,fill=gold]{2pt} & $31.10$ \tikzcircle[silver,fill=silver]{2pt} \\
TensoRF~\cite{chen2022tensorf} & $33.14$ & $35.76$ \tikzcircle[bronze,fill=bronze]{2pt} & $26.01$ & $33.99$ & $37.41$ \tikzcircle[bronze,fill=bronze]{2pt} & $36.46$ \tikzcircle[bronze,fill=bronze]{2pt} & $30.12$ \tikzcircle[bronze,fill=bronze]{2pt} & $34.61$ & $30.77$ \\
Factor Fields~\cite{chen2023factor} & $33.14$ & - & - & - & - & - & - & - & - \\
K-Planes~\cite{fridovich2023k} & $32.36$ & $34.99$ & $25.66$ & $31.41$ & $36.78$ & $35.75$ & $29.48$ & $34.05$ & $30.74$ \\
\midrule
Ours & $34.62$ \tikzcircle[gold,fill=gold]{2pt} & $36.74$ \tikzcircle[gold,fill=gold]{2pt} & $26.47$ \tikzcircle[gold,fill=gold]{2pt} & $35.14$ \tikzcircle[silver,fill=silver]{2pt} & $38.65$ \tikzcircle[gold,fill=gold]{2pt} & $37.53$ \tikzcircle[gold,fill=gold]{2pt} & $34.30$ \tikzcircle[silver,fill=silver]{2pt} & $36.17$ \tikzcircle[silver,fill=silver]{2pt} & $31.94$ \tikzcircle[gold,fill=gold]{2pt} \\
Ours$_{3.66M}$ & $33.97$ \tikzcircle[silver,fill=silver]{2pt} & $35.82$ \tikzcircle[silver,fill=silver]{2pt} & $26.19$ \tikzcircle[silver,fill=silver]{2pt} & $34.08$ \tikzcircle[bronze,fill=bronze]{2pt} & $38.11$ \tikzcircle[silver,fill=silver]{2pt} & $36.75$ \tikzcircle[silver,fill=silver]{2pt} & $34.32$ \tikzcircle[gold,fill=gold]{2pt} & $35.49$ & $31.03$ \tikzcircle[bronze,fill=bronze]{2pt} \\
\midrule
\midrule
\multicolumn{2}{l}{\textbf{SSIM$\uparrow$}} \\
\midrule
NeRF~\cite{mildenhall2021nerf} & $0.947$ & $0.967$ & $0.925$ & $0.964$ & $0.974$ & $0.961$ & $0.949$ & $0.980$ & $0.856$ \\
Mip-NeRF 360~\cite{barron2022mip} & $0.962$ & - & - & - & - & - & - & - & - \\
Point-NeRF~\cite{xu2022point} & $0.962$ & $0.984$ \tikzcircle[bronze,fill=bronze]{2pt} & $0.935$ & $0.987$ \tikzcircle[gold,fill=gold]{2pt} & $0.982$ \tikzcircle[bronze,fill=bronze]{2pt} & $0.978$ & $0.948$ & $0.990$ \tikzcircle[silver,fill=silver]{2pt} & $0.892$ \\
Plenoxels~\cite{fridovich2022plenoxels} & $0.958$ & $0.977$ & $0.933$ & $0.976$ & $0.980$ & $0.976$ & $0.949$ & $0.985$ & $0.890$ \\
Instant NGP~\cite{muller2022instant} & $0.963$ \tikzcircle[bronze,fill=bronze]{2pt} & $0.985$ \tikzcircle[silver,fill=silver]{2pt} & $0.940$ \tikzcircle[bronze,fill=bronze]{2pt} & $0.982$ \tikzcircle[bronze,fill=bronze]{2pt} & $0.982$ \tikzcircle[bronze,fill=bronze]{2pt} & $0.982$ & $0.949$ & $0.989$ \tikzcircle[bronze,fill=bronze]{2pt} & $0.893$ \\
TensoRF~\cite{chen2022tensorf} & $0.963$ \tikzcircle[bronze,fill=bronze]{2pt} & $0.985$ \tikzcircle[silver,fill=silver]{2pt} & $0.937$ & $0.982$ \tikzcircle[bronze,fill=bronze]{2pt} & $0.982$ \tikzcircle[bronze,fill=bronze]{2pt} & $0.983$ \tikzcircle[bronze,fill=bronze]{2pt} & $0.952$ \tikzcircle[silver,fill=silver]{2pt} & $0.988$ & $0.895$ \\
Factor Fields~\cite{chen2023factor} & $0.961$ & - & - & - & - & - & - & - & - \\
K-Planes~\cite{fridovich2023k} & $0.962$ & $0.983$ & $0.938$ & $0.975$ & $0.982$ \tikzcircle[bronze,fill=bronze]{2pt} & $0.982$ & $0.950$ \tikzcircle[bronze,fill=bronze]{2pt} & $0.988$ & $0.897$ \tikzcircle[bronze,fill=bronze]{2pt} \\
\midrule
Ours & $0.975$ \tikzcircle[gold,fill=gold]{2pt} & $0.988$ \tikzcircle[gold,fill=gold]{2pt} & $0.946$ \tikzcircle[gold,fill=gold]{2pt} & $0.987$ \tikzcircle[gold,fill=gold]{2pt} & $0.987$ \tikzcircle[gold,fill=gold]{2pt} & $0.986$ \tikzcircle[gold,fill=gold]{2pt} & $0.980$ \tikzcircle[gold,fill=gold]{2pt} & $0.992$ \tikzcircle[gold,fill=gold]{2pt} & $0.930$ \tikzcircle[gold,fill=gold]{2pt} \\
Ours$_{3.66M}$ & $0.971$ \tikzcircle[silver,fill=silver]{2pt} & $0.985$ \tikzcircle[silver,fill=silver]{2pt} & $0.942$ \tikzcircle[silver,fill=silver]{2pt} & $0.984$ \tikzcircle[silver,fill=silver]{2pt} & $0.985$ \tikzcircle[silver,fill=silver]{2pt} & $0.984$ \tikzcircle[silver,fill=silver]{2pt} & $0.980$ \tikzcircle[gold,fill=gold]{2pt} & $0.990$ \tikzcircle[silver,fill=silver]{2pt} & $0.919$ \tikzcircle[silver,fill=silver]{2pt} \\
\midrule
\midrule
\multicolumn{2}{l}{\textbf{LPIPS$_{VGG}$$\downarrow$}} \\
\midrule
NeRF~\cite{mildenhall2021nerf} & $0.081$ & $0.046$ & $0.091$ & $0.044$ & $0.121$ & $0.050$ & $0.063$ & $0.028$ & $0.206$ \\
Mip-NeRF 360~\cite{barron2022mip} & $0.039$ \tikzcircle[silver,fill=silver]{2pt} & - & - & - & - & - & - & - & - \\
Point-NeRF~\cite{xu2022point} & $0.050$ & $0.023$ & $0.078$ & $0.022$ \tikzcircle[bronze,fill=bronze]{2pt} & $0.037$ & $0.024$ & $0.072$ & $0.014$ \tikzcircle[bronze,fill=bronze]{2pt} & $0.124$ \tikzcircle[silver,fill=silver]{2pt} \\
Plenoxels~\cite{fridovich2022plenoxels} & $0.049$ & $0.031$ & $0.067$ \tikzcircle[bronze,fill=bronze]{2pt} & $0.026$ & $0.037$ & $0.028$ & $0.057$ \tikzcircle[bronze,fill=bronze]{2pt} & $0.015$ & $0.134$ \tikzcircle[bronze,fill=bronze]{2pt} \\
Instant NGP~\cite{muller2022instant} & $0.051$ & $0.023$ & $0.076$ & $0.027$ & $0.038$ & $0.021$ \tikzcircle[bronze,fill=bronze]{2pt} & $0.065$ & $0.020$ & $0.137$ \\
TensoRF~\cite{chen2022tensorf} & $0.047$ \tikzcircle[bronze,fill=bronze]{2pt} & $0.022$ \tikzcircle[bronze,fill=bronze]{2pt} & $0.073$ & $0.022$ \tikzcircle[bronze,fill=bronze]{2pt} & $0.032$ \tikzcircle[bronze,fill=bronze]{2pt} & $0.018$ \tikzcircle[silver,fill=silver]{2pt} & $0.058$ & $0.015$ & $0.138$ \\
Factor Fields~\cite{chen2023factor} & - & - & - & - & - & - & - & - & - \\
K-Planes~\cite{fridovich2023k} & $0.062$ & $0.027$ & $0.089$ & $0.056$ & $0.034$ & $0.047$ & $0.068$ & $0.029$ & $0.148$ \\
\midrule
Ours & $0.034$ \tikzcircle[gold,fill=gold]{2pt} & $0.015$ \tikzcircle[gold,fill=gold]{2pt} & $0.059$ \tikzcircle[gold,fill=gold]{2pt} & $0.014$ \tikzcircle[gold,fill=gold]{2pt} & $0.021$ \tikzcircle[gold,fill=gold]{2pt} & $0.015$ \tikzcircle[gold,fill=gold]{2pt} & $0.031$ \tikzcircle[gold,fill=gold]{2pt} & $0.008$ \tikzcircle[gold,fill=gold]{2pt} & $0.110$ \tikzcircle[gold,fill=gold]{2pt} \\
Ours$_{3.66M}$ & $0.039$ \tikzcircle[silver,fill=silver]{2pt} & $0.019$ \tikzcircle[silver,fill=silver]{2pt} & $0.065$ \tikzcircle[silver,fill=silver]{2pt} & $0.019$ \tikzcircle[silver,fill=silver]{2pt} & $0.025$ \tikzcircle[silver,fill=silver]{2pt} & $0.018$ \tikzcircle[silver,fill=silver]{2pt} & $0.034$ \tikzcircle[silver,fill=silver]{2pt} & $0.010$ \tikzcircle[silver,fill=silver]{2pt} & $0.124$ \tikzcircle[silver,fill=silver]{2pt} \\
\midrule
\midrule
\multicolumn{2}{l}{\textbf{LPIPS$_{Alex}$$\downarrow$}} \\
\midrule
Point-NeRF~\cite{xu2022point} & $0.028$ & $0.010$ & $0.055$ & $0.009$ \tikzcircle[silver,fill=silver]{2pt} & $0.016$ & $0.011$ & $0.041$ & $0.007$ \tikzcircle[bronze,fill=bronze]{2pt} & $0.070$ \tikzcircle[silver,fill=silver]{2pt} \\
Instant NGP~\cite{muller2022instant} & $0.028$ & $0.0097$ \tikzcircle[bronze,fill=bronze]{2pt} & $0.0540$ & $0.0174$ & $0.0142$ & $0.0085$ \tikzcircle[bronze,fill=bronze]{2pt} & $0.0296$ & $0.0072$ & $0.0863$ \\
TensoRF~\cite{chen2022tensorf} & $0.027$ \tikzcircle[bronze,fill=bronze]{2pt} & $0.010$ & $0.051$ \tikzcircle[bronze,fill=bronze]{2pt} & $0.012$ & $0.013$ \tikzcircle[bronze,fill=bronze]{2pt} & $0.007$ \tikzcircle[silver,fill=silver]{2pt} & $0.026$ \tikzcircle[bronze,fill=bronze]{2pt} & $0.009$ & $0.085$ \\
K-Planes~\cite{fridovich2023k} & $0.031$ & $0.0125$ & $0.0527$ & $0.0209$ & $0.0170$ & $0.0096$ & $0.0303$ & $0.0091$ & $0.0968$ \\
\midrule
Ours & $0.018$ \tikzcircle[gold,fill=gold]{2pt} & $0.0067$ \tikzcircle[gold,fill=gold]{2pt} & $0.0409$ \tikzcircle[gold,fill=gold]{2pt} & $0.0085$ \tikzcircle[gold,fill=gold]{2pt} & $0.0085$ \tikzcircle[gold,fill=gold]{2pt} & $0.0057$ \tikzcircle[gold,fill=gold]{2pt} & $0.0106$ \tikzcircle[gold,fill=gold]{2pt} & $0.0044$ \tikzcircle[gold,fill=gold]{2pt} & $0.0614$ \tikzcircle[gold,fill=gold]{2pt} \\
Ours$_{3.66M}$ & $0.022$ \tikzcircle[silver,fill=silver]{2pt} & $0.0088$ \tikzcircle[silver,fill=silver]{2pt} & $0.0454$ \tikzcircle[silver,fill=silver]{2pt} & $0.0101$ \tikzcircle[bronze,fill=bronze]{2pt} & $0.0109$ \tikzcircle[silver,fill=silver]{2pt} & $0.0070$ \tikzcircle[silver,fill=silver]{2pt} & $0.0119$ \tikzcircle[silver,fill=silver]{2pt} & $0.0060$ \tikzcircle[silver,fill=silver]{2pt} & $0.0735$ \tikzcircle[bronze,fill=bronze]{2pt} \\
\bottomrule
\end{tabular}
}
\end{center}
\caption{
\textbf{Neural Radiance Field Reconstruction.} Per-scene quantitative comparisons (PSNR$\uparrow$, SSIM$\uparrow$, LPIPS$_{VGG}$$\downarrow$, LPIPS$_{Alex}$$\downarrow$) on the Synthetic NeRF dataset~\cite{mildenhall2021nerf}. Best 3 scores in each scene are marked with gold \tikzcircle[gold,fill=gold]{2pt}, silver \tikzcircle[silver,fill=silver]{2pt} and bronze \tikzcircle[bronze,fill=bronze]{2pt}.
``-" denotes scores that are unavailable in prior work. For LPIPS$_{Alex}$, since the scores of NeRF~\cite{mildenhall2021nerf}, Mip-NeRF 360~\cite{barron2022mip}, Plenoxels~\cite{fridovich2022plenoxels} and Factor Fields~\cite{chen2023factor} are unavailable in prior work, we exclude these methods in this metric.
}
\label{tab:nerf_syn_perscene}
\end{table*}
\clearpage

\clearpage
\clearpage

\clearpage
\clearpage

\clearpage
\begin{table*}[t]
\begin{center}
\begin{tabular}{l|l|llllllll}
\toprule
Methods & Avg. & Room & Fern & Leaves & Fortress & Orchids & Flower & T-Rex & Horns \\
\midrule
\midrule
\multicolumn{2}{l}{\textbf{PSNR$\uparrow$}} \\
\midrule
NeRF~\cite{mildenhall2021nerf} & $26.50$ & $32.70$ \tikzcircle[silver,fill=silver]{2pt} & $25.17$ & $20.92$ & $31.16$ \tikzcircle[silver,fill=silver]{2pt} & $20.36$ \tikzcircle[gold,fill=gold]{2pt} & $27.40$ & $26.80$ & $27.45$ \\
Mip-NeRF 360~\cite{barron2022mip} & $26.86$ \tikzcircle[bronze,fill=bronze]{2pt} & - & - & - & - & - & - & - & - \\
Plenoxels~\cite{fridovich2022plenoxels} & $26.29$ & $30.22$ & $25.46$ \tikzcircle[silver,fill=silver]{2pt} & $21.41$ \tikzcircle[silver,fill=silver]{2pt} & $31.09$ \tikzcircle[bronze,fill=bronze]{2pt} & $20.24$ \tikzcircle[bronze,fill=bronze]{2pt} & $27.83$ \tikzcircle[silver,fill=silver]{2pt} & $26.48$ & $27.58$ \\
TensoRF~\cite{chen2022tensorf} & $26.73$ & $32.35$ & $25.27$ & $21.30$ \tikzcircle[bronze,fill=bronze]{2pt} & $31.36$ \tikzcircle[gold,fill=gold]{2pt} & $19.87$ & $28.60$ \tikzcircle[bronze,fill=bronze]{2pt} & $26.97$ \tikzcircle[bronze,fill=bronze]{2pt} & $28.14$ \tikzcircle[bronze,fill=bronze]{2pt} \\
K-Planes~\cite{fridovich2023k} & $26.92$ \tikzcircle[silver,fill=silver]{2pt} & $32.64$ \tikzcircle[bronze,fill=bronze]{2pt} & $25.38$ \tikzcircle[bronze,fill=bronze]{2pt} & $21.30$ \tikzcircle[bronze,fill=bronze]{2pt} & $30.44$ & $20.26$ \tikzcircle[silver,fill=silver]{2pt} & $28.67$ \tikzcircle[gold,fill=gold]{2pt} & $28.01$ \tikzcircle[silver,fill=silver]{2pt} & $28.64$ \tikzcircle[silver,fill=silver]{2pt} \\
\midrule
Ours & $27.05$ \tikzcircle[gold,fill=gold]{2pt} & $32.80$ \tikzcircle[gold,fill=gold]{2pt} & $25.48$ \tikzcircle[gold,fill=gold]{2pt} & $21.81$ \tikzcircle[gold,fill=gold]{2pt} & $30.98$ & $20.03$ & $28.57$ & $28.06$ \tikzcircle[gold,fill=gold]{2pt} & $28.68$ \tikzcircle[gold,fill=gold]{2pt} \\
\midrule
\midrule
\multicolumn{2}{l}{\textbf{SSIM$\uparrow$}} \\
\midrule
NeRF~\cite{mildenhall2021nerf} & $0.811$ & $0.948$ & 0.$792$ & $0.690$ & $0.881$ & $0.641$ & $0.827$ & $0.880$ & $0.828$ \\
Mip-NeRF 360~\cite{barron2022mip} & $0.858$ \tikzcircle[gold,fill=gold]{2pt} & - & - & - & - & - & - & - & - \\
Plenoxels~\cite{fridovich2022plenoxels} & $0.839$ & $0.937$ & $0.832$ \tikzcircle[gold,fill=gold]{2pt} & $0.760$ \tikzcircle[silver,fill=silver]{2pt} & $0.885$ & $0.687$ \tikzcircle[gold,fill=gold]{2pt} & $0.862$ & $0.890$ & $0.857$ \\
TensoRF~\cite{chen2022tensorf} & $0.839$ & $0.952$ \tikzcircle[bronze,fill=bronze]{2pt} & $0.814$ & $0.752$ \tikzcircle[bronze,fill=bronze]{2pt} & $0.897$ \tikzcircle[gold,fill=gold]{2pt} & $0.649$ & $0.871$ \tikzcircle[silver,fill=silver]{2pt} & $0.900$ \tikzcircle[bronze,fill=bronze]{2pt} & $0.877$ \tikzcircle[bronze,fill=bronze]{2pt} \\
K-Planes~\cite{fridovich2023k} & $0.847$ \tikzcircle[bronze,fill=bronze]{2pt} & $0.957$ \tikzcircle[gold,fill=gold]{2pt} & $0.828$ \tikzcircle[silver,fill=silver]{2pt} & $0.746$ & $0.890$ \tikzcircle[bronze,fill=bronze]{2pt} & $0.676$ \tikzcircle[silver,fill=silver]{2pt} & $0.872$ \tikzcircle[gold,fill=gold]{2pt} & $0.915$ \tikzcircle[silver,fill=silver]{2pt} & $0.892$ \tikzcircle[silver,fill=silver]{2pt} \\
\midrule
Ours & $0.849$ \tikzcircle[silver,fill=silver]{2pt} & $0.955$ \tikzcircle[silver,fill=silver]{2pt} & $0.822$ \tikzcircle[bronze,fill=bronze]{2pt} & $0.769$ \tikzcircle[gold,fill=gold]{2pt} & $0.891$ \tikzcircle[silver,fill=silver]{2pt} & $0.675$ \tikzcircle[bronze,fill=bronze]{2pt} & $0.868$ \tikzcircle[bronze,fill=bronze]{2pt} & $0.916$ \tikzcircle[gold,fill=gold]{2pt} & $0.895$ \tikzcircle[gold,fill=gold]{2pt} \\
\midrule
\midrule
\multicolumn{2}{l}{\textbf{LPIPS$_{VGG}$$\downarrow$}} \\
\midrule
NeRF~\cite{mildenhall2021nerf} & $0.250$ & $0.178$ & $0.280$ & $0.316$ & $0.171$ & $0.321$ & $0.219$ & $0.249$ & $0.268$ \\
Plenoxels~\cite{fridovich2022plenoxels} & $0.210$ & $0.192$ & $0.224$ \tikzcircle[bronze,fill=bronze]{2pt} & $0.198$ \tikzcircle[gold,fill=gold]{2pt} & $0.180$ & $0.242$ \tikzcircle[gold,fill=gold]{2pt} & $0.179$ & $0.238$ & $0.231$ \\
TensoRF~\cite{chen2022tensorf} & $0.204$ \tikzcircle[bronze,fill=bronze]{2pt} & $0.167$ \tikzcircle[bronze,fill=bronze]{2pt} & $0.237$ & $0.217$ \tikzcircle[silver,fill=silver]{2pt} & $0.148$ \tikzcircle[silver,fill=silver]{2pt} & $0.278$ & $0.169$ \tikzcircle[bronze,fill=bronze]{2pt} & $0.221$ \tikzcircle[bronze,fill=bronze]{2pt} & $0.196$ \tikzcircle[bronze,fill=bronze]{2pt} \\
K-Planes~\cite{fridovich2023k} & $0.194$ \tikzcircle[silver,fill=silver]{2pt} & $0.147$ \tikzcircle[silver,fill=silver]{2pt} & $0.223$ \tikzcircle[silver,fill=silver]{2pt} & $0.242$ & $0.154$ \tikzcircle[bronze,fill=bronze]{2pt} & $0.250$ \tikzcircle[silver,fill=silver]{2pt} & $0.165$ \tikzcircle[silver,fill=silver]{2pt} & $0.199$ \tikzcircle[silver,fill=silver]{2pt} & $0.173$ \tikzcircle[silver,fill=silver]{2pt} \\
\midrule
Ours & $0.179$ \tikzcircle[gold,fill=gold]{2pt} & $0.134$ \tikzcircle[gold,fill=gold]{2pt} & $0.209$ \tikzcircle[gold,fill=gold]{2pt} & $0.238$ \tikzcircle[bronze,fill=bronze]{2pt} & $0.128$ \tikzcircle[gold,fill=gold]{2pt} & $0.271$ \tikzcircle[bronze,fill=bronze]{2pt} & $0.147$ \tikzcircle[gold,fill=gold]{2pt} & $0.158$ \tikzcircle[gold,fill=gold]{2pt} & $0.149$ \tikzcircle[gold,fill=gold]{2pt} \\
\midrule
\midrule
\multicolumn{2}{l}{\textbf{LPIPS$_{Alex}$$\downarrow$}} \\
\midrule
TensoRF~\cite{chen2022tensorf} & $0.124$ \tikzcircle[bronze,fill=bronze]{2pt} & 0.082 \tikzcircle[bronze,fill=bronze]{2pt} & 0.155 \tikzcircle[bronze,fill=bronze]{2pt} & 0.153 \tikzcircle[silver,fill=silver]{2pt} & 0.075 \tikzcircle[bronze,fill=bronze]{2pt} & 0.201 \tikzcircle[bronze,fill=bronze]{2pt} & 0.106 \tikzcircle[bronze,fill=bronze]{2pt} & 0.099 \tikzcircle[bronze,fill=bronze]{2pt} & 0.123 \tikzcircle[bronze,fill=bronze]{2pt} \\
K-Planes~\cite{fridovich2023k} & $0.102$ \tikzcircle[silver,fill=silver]{2pt} & 0.066 \tikzcircle[silver,fill=silver]{2pt} & 0.130 \tikzcircle[silver,fill=silver]{2pt} & 0.153 \tikzcircle[silver,fill=silver]{2pt} & 0.068 \tikzcircle[silver,fill=silver]{2pt} & 0.151 \tikzcircle[gold,fill=gold]{2pt} & 0.088 \tikzcircle[silver,fill=silver]{2pt} & 0.071 \tikzcircle[silver,fill=silver]{2pt} & 0.092 \tikzcircle[silver,fill=silver]{2pt} \\
\midrule
Ours & $0.090$ \tikzcircle[gold,fill=gold]{2pt} & 0.059 \tikzcircle[gold,fill=gold]{2pt} & 0.111 \tikzcircle[gold,fill=gold]{2pt} & 0.127 \tikzcircle[gold,fill=gold]{2pt} & 0.056 \tikzcircle[gold,fill=gold]{2pt} & 0.160 \tikzcircle[silver,fill=silver]{2pt} & 0.072 \tikzcircle[gold,fill=gold]{2pt} & 0.057 \tikzcircle[gold,fill=gold]{2pt} & 0.075 \tikzcircle[gold,fill=gold]{2pt} \\
\bottomrule
\end{tabular}
\end{center}
\caption{
\textbf{Neural Radiance Field Reconstruction.} Per-scene quantitative comparisons (PSNR$\uparrow$, SSIM$\uparrow$, LPIPS$_{VGG}$$\downarrow$, LPIPS$_{Alex}$$\downarrow$) on the real LLFF Forward-Facing dataset~\cite{mildenhall2019local}. Best 3 scores in each scene are marked with gold \tikzcircle[gold,fill=gold]{2pt}, silver \tikzcircle[silver,fill=silver]{2pt} and bronze \tikzcircle[bronze,fill=bronze]{2pt}.
``-" denotes scores that are unavailable in prior work.
For LPIPS$_{Alex}$, since the scores of NeRF~\cite{mildenhall2021nerf}, Mip-NeRF 360~\cite{barron2022mip} and Plenoxels~\cite{fridovich2022plenoxels} are unavailable in prior work, we exclude these methods in this metric.
}
\label{tab:nerf_llff_perscene}
\end{table*}
\clearpage

\clearpage

\end{document}